\documentclass[10pt,twocolumn,letterpaper]{article}

\usepackage{cvpr}              

\usepackage{xcolor}
\usepackage{amsmath}
\usepackage{amssymb}
\usepackage{array}
\usepackage{multirow}
\usepackage{color}
\usepackage{colortbl}
\usepackage{framed}
\usepackage{bm}
\usepackage{bbm}
\usepackage{xspace}
\usepackage{enumitem}
\usepackage{algorithm}
\usepackage{listings}
\usepackage{url}
\usepackage{booktabs}
\usepackage{pifont} 
\usepackage[accsupp]{axessibility}
\usepackage{lipsum}
\usepackage{bm}
\definecolor{LightCyan}{rgb}{0.88,0.95,1}
\definecolor{blond}{rgb}{0.98, 0.94, 0.75}
\definecolor{green}{rgb}{0.0, 0.62, 0.42}
\definecolor{LightGray}{gray}{0.93}
\definecolor{ourcolor}{rgb}{0.925, 0.902, 0.969}

\newcommand{\cmark}{\ding{51}}%
\newcommand{\xmark}{\ding{55}}%

\newcommand{\ours}{ReT\xspace}

\def \ie {\emph{i.e.}}
\def \eg {\emph{e.g.}}

\newcommand{\tit}[1]{\smallbreak\noindent\textbf{#1.}}
\newcommand{\tinytit}[1]{\noindent\textbf{#1.}}

\newcommand{\query}{\operatorname{\bm{q}}}
\newcommand{\doc}{\operatorname{\bm{d}}}
\newcommand{\retQ}{\operatorname{\text{\ours}_{\textbf{Q}}}}
\newcommand{\retD}{\operatorname{\text{\ours}_{\textbf{D}}}}

\newcommand{\inputH}{\bm{h}_l}
\newcommand{\nextH}{\bm{h}_{l+1}}
\newcommand{\finalH}{\bm{h}_{L}}
\newcommand{\finalHhat}{\overline{\bm{h}}_{L}}
\newcommand{\hatH}{\hat{\bm{h}_l}}

\newcommand{\candidate}{\mathbf{c}_l}
\newcommand{\candidateNext}{\bm{c}_{l+1}}

\newcommand{\genericZ}{\bm{z}^m_l}
\newcommand{\textZ}{\bm{z}^T_l}
\newcommand{\visZ}{\bm{z}^V_l}

\newcommand{\textInputGate}{\bm{i}^T_l}
\newcommand{\visInputGate}{\bm{i}^V_l}
\newcommand{\genericInputGate}{\bm{i}^m_l}
\newcommand{\forgetGate}{\bm{f}_l}

\newcommand{\genericInputW}{\bm{W}^m_i}
\newcommand{\textForgetW}{\bm{W}^T_f}
\newcommand{\visForgetW}{\bm{W}^V_f}

\definecolor{cvprblue}{rgb}{0.21,0.49,0.74}
\usepackage[pagebackref,breaklinks,colorlinks,allcolors=cvprblue]{hyperref}
\usepackage{graphicx}%
\usepackage{multirow}%
\usepackage{amsthm}%
\usepackage{mathrsfs}%
\usepackage[title]{appendix}%
\usepackage{xcolor}%
\usepackage{textcomp}%
\usepackage{booktabs}%
\usepackage{listings}%
\usepackage{array}
\usepackage{amsmath,amsfonts}
\usepackage{stfloats}
\usepackage{url}
\usepackage{verbatim}
\usepackage{bm}
\usepackage{bbm}
\usepackage{xspace}
\usepackage{enumitem}
\usepackage{pifont}
\usepackage{color}
\usepackage{colortbl}
\usepackage{ctable} 
\usepackage{xcolor}%

\usepackage{bbding}
\usepackage{fontawesome}

\def\paperID{9860} 
\def\confName{CVPR}
\def\confYear{2025}

\title{Recurrence-Enhanced Vision-and-Language Transformers\\ for Robust Multimodal Document Retrieval}

\author{Davide Caffagni$^1$\thanks{Equal contribution.} \quad Sara Sarto$^1$\footnotemark[1] \quad Marcella Cornia$^1$ \quad Lorenzo Baraldi$^1$ \quad Rita  Cucchiara$^{1,2}$  \\
$^1$University of Modena and Reggio Emilia, Modena, Italy \quad $^2$IIT-CNR, Pisa, Italy\\
{\tt\small \{name.surname\}@unimore.it}
}

\begin{document}
\maketitle
\begin{abstract}
Cross-modal retrieval is gaining increasing efficacy and interest from the research community, thanks to large-scale training, novel architectural and learning designs, and its application in LLMs and multimodal LLMs. In this paper, we move a step forward and design an approach that allows for multimodal queries -- composed of both an image and a text -- and can search within collections of multimodal documents, where images and text are interleaved. Our model, \ours, employs multi-level representations extracted from different layers of both visual and textual backbones, both at the query and document side. To allow for multi-level and cross-modal understanding and feature extraction, \ours employs a novel Transformer-based recurrent cell that integrates both textual and visual features at different layers, and leverages sigmoidal gates inspired by the classical design of LSTMs. Extensive experiments on M2KR and M-BEIR benchmarks show that \ours achieves state-of-the-art performance across diverse settings.
Our source code and trained models are publicly available at: {\url{https://github.com/aimagelab/ReT}}.
\end{abstract}    
\section{Introduction}
\label{sec:intro}
Retrieving useful information to satisfy an input query is a fundamental problem in Computer Vision and Multimedia. In early years, retrieval was done mostly in unimodal settings, where a query is expressed either with an image or in natural language and the retrievable set of items belongs to the same modality~\cite{izacard2021unsupervised,zheng2017sift,noh2017large,radenovic2018fine}. Recently, we have been experiencing a progressive shift towards multimodal data~\cite{lin2014microsoft,sharma2018conceptual,schuhmann2022laion}. Real-world applications currently require at least to understand natural language queries to retrieve images, or vice versa~\cite{radford2021learning,sarto2024towards}. This trend is expected to give rise to the need for retrieval engines capable of handling more data modalities, such as video and audio~\cite{girdhar2023imagebind}.

\begin{figure}[t]
\vspace{-0.15cm}
    \centering
    \includegraphics[width=0.97\linewidth]{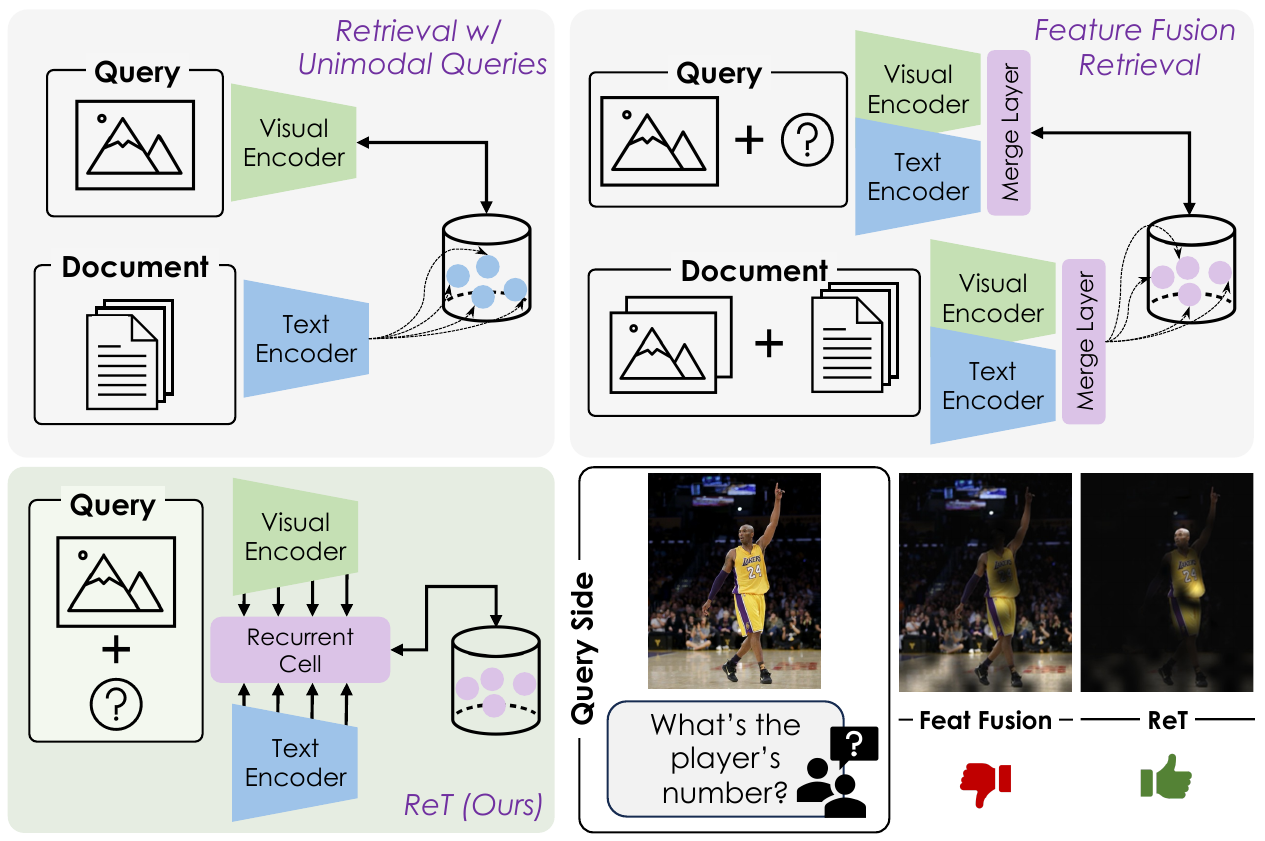}
    \vspace{-0.15cm}
    \caption{
    Comparison between cross-modal retrieval with unimodal queries (top left), retrieval with multimodal queries with feature fusion~\cite{wei2024uniir} (top right), and our \ours (bottom left). Our approach enables retrieval with multimodal queries employing a recurrent, multi-level feature extraction process.
    }
    \label{fig:first_page}
    \vspace{-0.4cm}
\end{figure}

Concurrently, there is a growing need to give retrieval systems the ability to deal with \textit{multimodal queries} and the capacity of searching among \textit{multimodal documents}, \ie~in settings in which multiple modalities coexist both inside the query and inside the retrievable items. One of such cases is that of queries consisting of an image paired with a text that poses a question related to the image or specifies what in the image should be retrieved~\cite{wei2024uniir,lin2024preflmr,chen2023can,mensink2023encyclopedic}. Similarly, the latter is the case of multimodal documents, in which text and images are naturally interleaved. The current state of the art for multimodal retrieval~\cite{radford2021learning,li2023blip,zhai2023sigmoid} falls short in accommodating this flexibility in user demands, as it mainly supports single-modality queries and retrievable documents.

In an effort to address these needs, in this paper, we propose a retrieval approach that natively supports multimodal (\ie, composed of images and text) queries and retrievable items. Differently from existing proposals in the field~\cite{wei2024uniir}, that fuse feature extracted from the last layer of vision-and-language backbones, our proposal employs multi-layer representations in both modalities, while still allowing a late interaction schema (Fig.~\ref{fig:first_page}). Indeed, we posit that including representations from shallower layers of vision-and-language models can help to address the wide variety of multimodal queries and documents. For instance, an existing retrieval model can naturally recognize an image of a famous basketball player, such as Kobe Bryant. However, the number on its jersey is a low-level detail that may be lost in the bottom-up information stream of its image encoder, which is biased toward the main entity in the foreground. On the other hand, shallow layers do naturally hold feature representations of low-level visual clues.

As an additional ingredient, in contrast to standard retrievers that compute query-document relevance score as the cosine similarity between a single latent vector generated from a query and a document, our model encodes every query and document with multiple tokens that are used for late interaction. This way, the comparisons between each query token and each document token can capture fine-grained local similarities that are aggregated to obtain a more robust final score~\cite{santhanam2022colbertv2, lin2023fine, lin2024preflmr}.

To achieve the aforementioned goals, we design a novel Transformer-based recurrent cell, that for each layer merges the features from the visual backbone and the textual backbone with its internal states. Inspired by the gating mechanism of an LSTM~\cite{hochreiter1997long}, our model features a forget gate that controls how much information to retain from shallower layers. On the other hand, a textual and a visual input gate modulates the unimodal information flow for each layer of the vision-and-language backbones. This allows \ours to decide on which layer focus the representation on, and within each layer, which modality is the most needed for encoding.

Our model, which we name \textbf{\ours} (short for \textit{Recurrence-enhanced Transformer} and for \textit{Retrieval}) is experimentally validated on the challenging M2KR~\cite{lin2024preflmr} benchmark, which features a collection of diverse datasets converted for multimodal retrieval. To further extend the evaluation, we augment the OVEN~\cite{hu2023open}, InfoSeek~\cite{chen2023can}, Encyclopedic-VQA~\cite{mensink2023encyclopedic}, and OKVQA~\cite{marino2019ok} splits of M2KR to also include images on the reference documents to be retrieved. Experimentally, \ours outscores the current state-of-the-art method on most datasets of the M2KR suite. Additionally, we demonstrate its effectiveness when tested on other multimodal retrieval benchmarks like M-BEIR~\cite{wei2024uniir} and when employed as the retrieval component in a retrieval-augmented generation pipeline. Our source code and trained models will be made publicly available.
\section{Related Work}
\label{sec:related}
\tinytit{Information Retrieval}
Information retrieval has advanced significantly in the multimodal domain, especially in image-to-text matching. Modern retrieval systems now integrate visual and textual data to address complex, real-world queries across both modalities.
Traditional image-text retrieval approaches mainly relied on dual-encoder architectures, like CLIP~\cite{radford2021learning}, ALIGN~\cite{jia2021scaling}, and other derivative variants~\cite{zhai2023sigmoid,sun2023eva}. These models use contrastive learning to align visual and textual representations, leveraging image-text pairs to train networks for cross-modal search. While impactful, these models are typically benchmarked on popular crowd-sourced datasets, including Flickr8k~\cite{hodosh2013framing}, Flickr30k~\cite{plummer2015flickr30k}, and COCO~\cite{lin2014microsoft}. Their limited size of these datasets, and their reliance on unimodal queries, restricts the performance and generalization of retrieval models, presenting a challenge to further advancements in this field.

\tit{Multimodal Document Retrieval}
With a rising demand for multimodal retrieval systems, the ability to handle complex multimodal queries has become essential. This trend has led to the development of specialized benchmarks for multimodal retrieval, supporting diverse tasks and data configurations.
For example, M2KR~\cite{lin2024preflmr} comprises several datasets adapted for multimodal retrieval tasks, such as OKVQA~\cite{marino2019ok}, OVEN~\cite{hu2023open} and InfoSeek~\cite{chen2023can}.
Another recent large-scale retrieval benchmark, M-BEIR~\cite{wei2024uniir}, supports multimodal retrieval tasks and comprises a collection of datasets from a variety of domains and image sources.

Achieving high performance in complex multimodal retrieval tasks has driven recent research to focus on effective strategies for encoding multimodal queries and documents. The UniIR architecture~\cite{wei2024uniir} employs pre-trained models like CLIP~\cite{radford2021learning} and BLIP~\cite{li2022blip} to integrate modalities,  aiming at a unified retriever for diverse tasks. Parallel efforts, including models like FLMR~\cite{lin2023fine}, PreFLMR~\cite{lin2024preflmr}, and the earlier ColBERT~\cite{khattab2020colbert}, have explored a late-interaction paradigm. This technique encodes multimodal queries and text-only documents independently into sets of latent tokens, with relevance scores computed by aggregating similarities between each query-document token pair. Building on this, we improve the richness of representations by leveraging information from multiple encoder layers.

\tit{Recurrence-Augmented Transformers} Transformer architectures~\cite{vaswani2017attention} have demonstrated remarkable performance across a wide range of tasks, from natural language understanding~\cite{brown2020language,wolf2020transformers,chowdhery2023palm,touvron2023llama} to Computer Vision applications~\cite{dosovitskiy2021image,touvron2021training,zhai2022scaling,khan2022transformers}. However, the quadratic complexity of Transformers with respect to input length has motivated significant research into alternative designs. To address this, various approaches have introduced recurrent mechanisms within Transformer-based models, interleaving Transformer and recurrent neural network layers to balance attention with recurrence~\cite{lei2021attention,lei2017simple,bapna2018best}. Other methods, such as the R-Transformer~\cite{wang2019r}, incorporate local recurrent cells to allow for parallel computation. Expanding on these techniques, the Block-Recurrent Transformer~\cite{hutchins2022block} embeds recurrent features directly into the Transformer framework, drawing from LSTM cell dynamics~\cite{hochreiter1997long}. Differently from these works that mainly leverage recurrence for computational reasons, in this paper, we leverage recurrence as a means for allowing multi-layer feature processing, with the ultimate goal of increasing performance on retrieval tasks.

\section{Proposed Method}
\label{sec:method}

\begin{figure*}[t]
    \centering
    \includegraphics[width=0.99\linewidth]{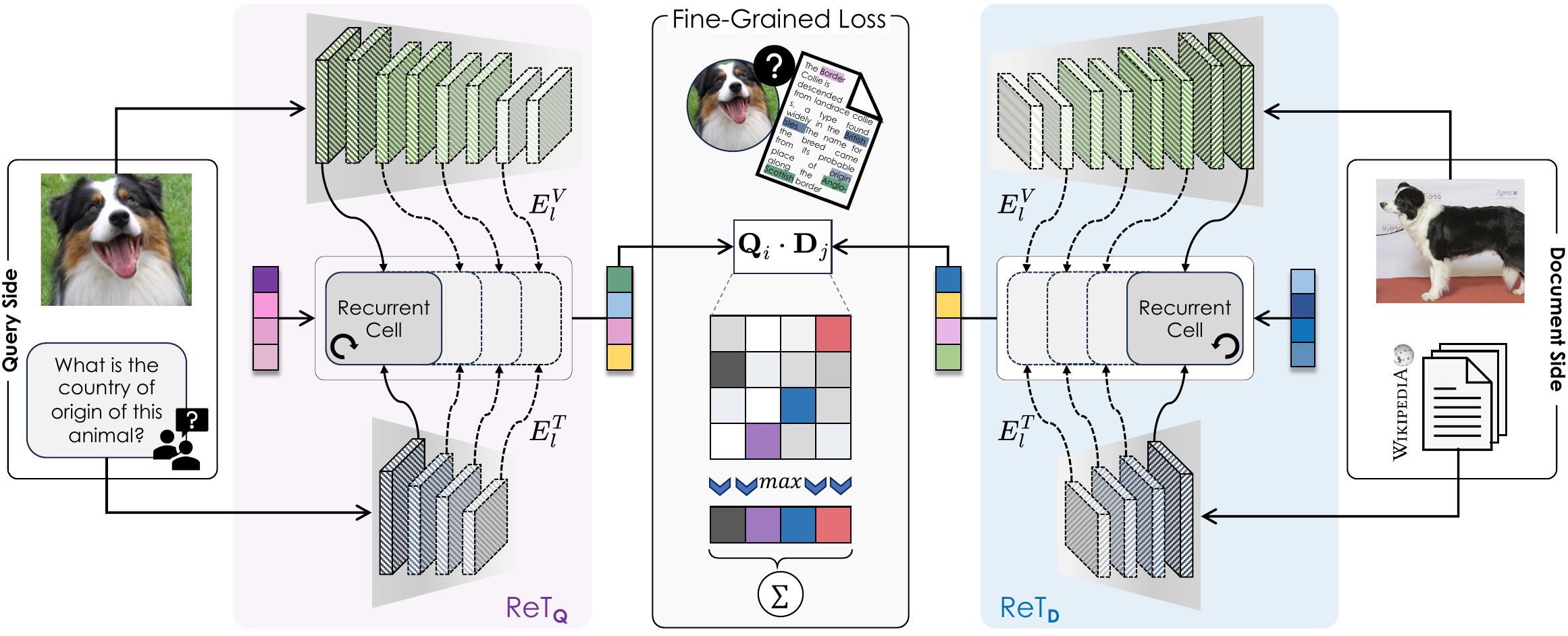}
    \vspace{-0.1cm}
    \caption{
    Overview of the proposed Recurrence-enhanced Transformer (\ours) for cross-modal retrieval with multimodal queries. Our model employs a Transformer-based recurrent cell to encode multiple vision-and-language layers into hidden vectors for similarity computation.
    }
    \label{fig:model}
    \vspace{-0.3cm}
\end{figure*}

In this section, we introduce our novel multimodal retriever model, termed \textbf{\ours}, to effectively search across heterogeneous data sources and retrieve information from a multimodal pool comprising a vast number of entities. 

\tit{Problem Setting}
In our setting, both queries $\query=(q^T, q^V)$ and documents $\doc=(d^T, d^V)$ are designed to represent paired image-text instances. Specifically, each query text $q^T$ consists of an instruction, \eg~``Utilizing the given image, obtain documents that respond to the following question'', followed by a query-specific question about the query image $q^V$ -- \eg~``What is the country of origin of this animal?''. 
The objective is to identify and retrieve documents that are mostly relevant to the query. 
Each document, similarly, consists of a textual component, $d^T$, that fulfils the query
and an optional accompanying image $d^V$.

\tit{Overview of the Approach}
Our proposed method, \ours, employs a novel Transformer-based recurrent cell to fuse multimodal features both on the query and document side, to compute a similarity score between queries and documents.
Features from textual and visual backbones are considered across multiple layers, thus providing an effective multi-level encoding of each modality composing queries and documents. At each recurrent step, \ours dynamically fuses its internal state with the visual and textual representations obtained from the current layer of the two backbones. Differently from a standard recurrent network applied to a temporal sequence, in \ours the concept of \textit{past} does not pertain to previous time steps, but rather to lower-level representations of a given query or document, obtained from shallower layers. A graphical overview illustrating the architecture of our model is reported in Fig.~\ref{fig:model}.

\tit{Learnable Gating}
Although being a Transformer-based architecture, our model manages the sequence of input layers through a learnable gating mechanism inspired by the LSTM architecture~\cite{hochreiter1997long}. This design can selectively forget lower-level representations in favor of higher-level features. Similarly, an input gate within each layer modulates the contribution of each modality, balancing the visual and textual inputs and providing a more controllable encoding.

\subsection{\ours: Enhancing Retrieval with Recurrence}
\label{sec:method_ret}
Our model consists of dedicated encoders for queries and documents, $\retQ$ and $\retD$, with identical architecture but distinct learnable weights, optimized jointly.
Given that  $\retQ$ and $\retD$ share the same architectural structure, for simplicity we describe only $\retQ$ in the following sections.

In particular, the $\retQ$ encoder consists of a recurrent cell and a pre-trained visual and textual backbone. In our notation, we define $E^m(q^m)=\left\{E^m_l(q^m)\right\}_{l=1}^{L}$ the set of activations gathered from each layer $l$ of the unimodal backbone, where $E^m_l(q^m)\in \mathbb{R}^{N \times d}$. Here, $m \in \{T, V\}$ specifies the modality of the backbone, either textual $T$ or visual $V$, and $L$ represents the total number of layers. 
Also, we denote the self-attention operator~\cite{vaswani2017attention} applied to a real-valued matrix $\mathbf{x}$ as $\text{Attention}(\mathbf{x})$. For cross-attention between two matrices, $\mathbf{x}$ and $\mathbf{y}$, we use the notation $\text{Attention}(\mathbf{x}, \mathbf{y})$.

\begin{figure}[t]
    \centering
    \includegraphics[width=0.98\linewidth]{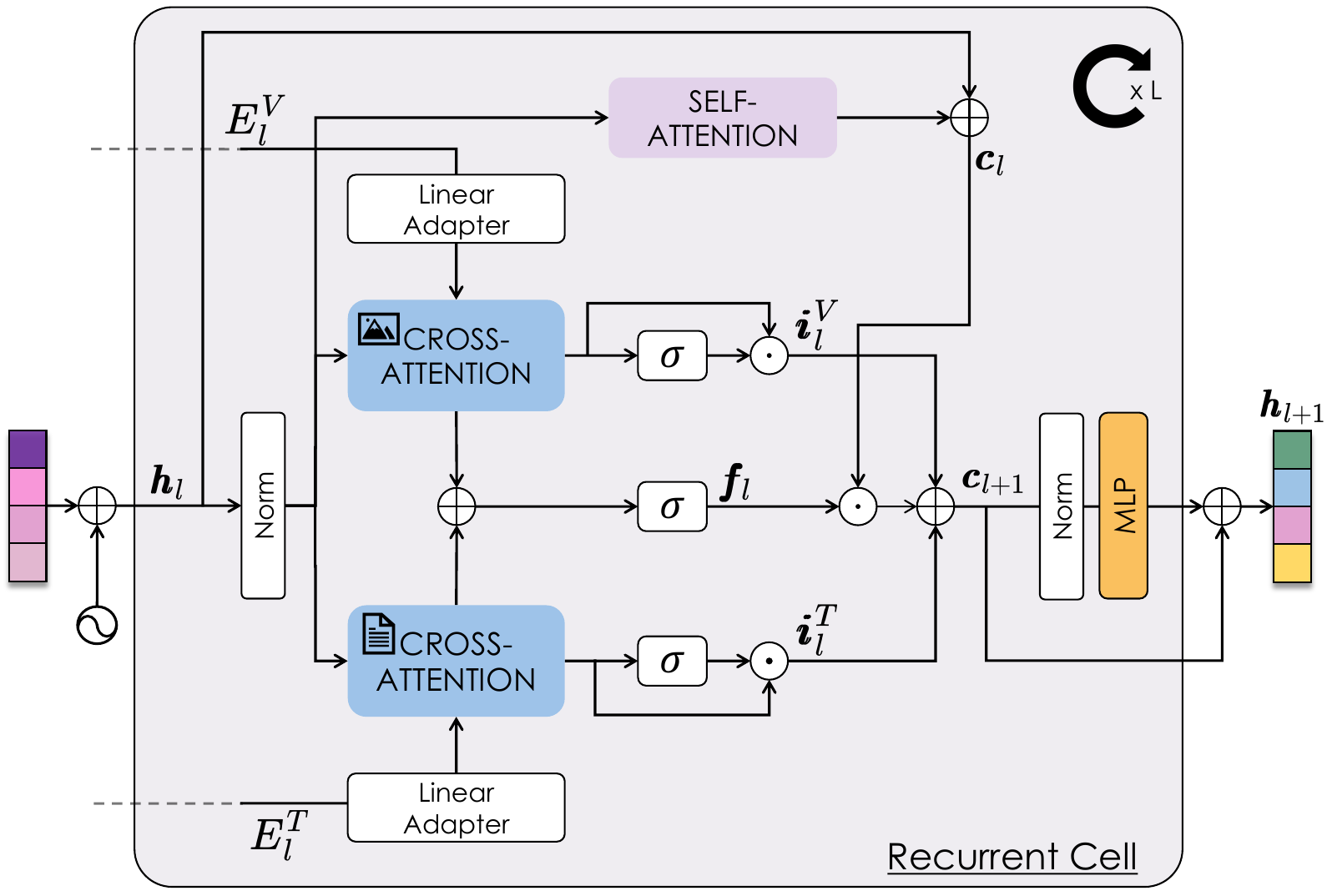}
    \vspace{-0.15cm}
    \caption{Graphical overview of the designed recurrent cell for the proposed retrieval model, which integrates layer-specific textual and visual features into a matricial hidden state.}
    \label{fig:recurrent_block}
    \vspace{-0.3cm}
\end{figure}

\tit{Recurrent Cells}
The architecture of the recurrent cell is illustrated in detail in Fig.~\ref{fig:recurrent_block}. At each layer $l$, it performs \textit{feature fusion}~\cite{wei2024uniir} across three different inputs: the output from the previous step $\mathbf{h}_l \in \mathbb{R}^{k \times d}$, the visual representation from the $l$-th layer of the visual backbone $\mathbf{E}^V_l$, and the textual representation from the $l$-th layer of the textual backbone $\mathbf{E}^T_l$. At the first iteration, the hidden state $\mathbf{h}_0$ is 
initialized to a set of $k$ learnable vectors.

Within the recurrent cell, following a layer normalization~\cite{lei2016layer}, the input $\inputH$ is fed to three parallel branches. The first branch computes the candidate state $\candidate$ of the recurrent cell, employing a self-attention operation with a residual connection to enhance stability and retain contextual information. Formally, the candidate state is computed as
\begin{equation}
    \label{eq:input}
    \candidate = \text{Attention}(\, \hatH \ ) + \inputH,
\end{equation}
where $\hatH = \texttt{LayerNorm}(\inputH)$. 
Given the permutation invariance of the attention operator, a fixed positional encoding~\cite{vaswani2017attention} is added to $\inputH$ before normalization. Notably, for $l \ge 1$, $\inputH$ will accumulate layer-specific encodings of the query image and of the text.

To incorporate contextual information from both modalities, two other parallel branches perform feature fusion with the unimodal representations from the $l$-th layers. Specifically, we employ two independent cross-attention modules to fuse the normalized input $\hatH$ with the visual and textual representations respectively, as
\begin{equation}
    \label{eq:unimodal_component}
    \genericZ =  \text{Attention}(\, \hatH, \, E^m_l \ ).
\end{equation} 
 
The three branches are finally combined to compute the new internal state of the recurrent cell, which is obtained as a linear combination of the candidate state $\candidate$ and the outputs from the two feature fusion branches $\textZ$ and $\visZ$, modulated by forget and input gates. Conceptually, the forget gate $\forgetGate$ determines how much information to retain from previous applications of the recurrent cell across the shallower layers (\ie, the ``past''), given the current multimodal information flow $\genericZ$. In contrast, the input gates $\genericInputGate$ control the contribution from the $l$-th layer of each unimodal backbone. This potentially shuts down interferences from high-level representations whenever the user query requires focusing on low-level details (\eg, colors and shapes).
Formally, the next candidate state is obtained as
\begin{equation}
    \label{eq:final}
    \candidateNext = \candidate \odot \forgetGate + \textZ \odot \textInputGate + \visZ \odot \visInputGate,
\end{equation}
where $\forgetGate$, $\textInputGate$ and $\visInputGate$ indicate the learnable sigmoidal gates. In particular, these are computed as follows:
\begin{equation}
\begin{aligned}
    \forgetGate &= \sigma(\textForgetW \cdot \textZ + \visForgetW \cdot \visZ + b_f), \\
    \genericInputGate &= \sigma(\genericInputW \cdot \genericZ + b_i),
\end{aligned}
\label{eq:forget_gate}
\end{equation}
where $\textForgetW, \visForgetW, \genericInputW$ are trainable weight matrices, and $b_f$, $b_i$ are fixed scalar biases.
The updated state $\candidateNext$ undergoes layer normalization and is passed through a residual two-layer feed-forward network to produce the output of the recurrent cell, as
\begin{equation}
    \label{eq:next_input}
    \nextH = \candidateNext +  \texttt{MLP}(\texttt{LayerNorm}(\candidateNext)).
\end{equation}

After going through different layers of the backbones, the output from the last iteration of the recurrent cell, $\finalH \in \mathbb{R}^{k \times d}$, consists of a set of latent tokens that serve to compute multiple query-document relevance scores. 
Specifically, the output $\finalH$ is transformed into a different vector space through a linear projection $\textbf{W}_{final} \in \mathbb{R}^{d \times \overline{d}}$, \ie 
\begin{equation}
    \label{eq:pre-output}
    \finalHhat = 
     \finalH \  \cdot \textbf{W}_{final} .
\end{equation}

\subsection{Training Procedure}
Given a query-document pair $(\query,\doc)$, along with a set of distinct learnable tokens in input for both the query and document sides, we denote the corresponding final output  $\finalHhat$ of the query and the document encoders as
\begin{align}
    \mathbf{Q} & =  \retQ(\query) \in \mathbb{R}^{k \times \overline{d}} \\
    \mathbf{D} & = \retD(\doc)^\intercal \in \mathbb{R}^{\overline{d} \times k}.
\end{align}

At training time, these representations are used to compute a fine-grained late-interaction~\cite{santhanam2022colbertv2} relevance score $s(\textbf{Q}, \textbf{D})$ between a query $\query$ and a document $\doc$, which is computed as a sum of maximum similarity values, \ie
\begin{equation}
    \label{eq:fine-grained loss}
    s(\textbf{Q}, \textbf{D}) = \sum_{i=1}^k \max_{j=1 \dots k} \textbf{Q}_i \cdot \textbf{D}_j.
\end{equation}

Notably, the similarity is implemented as the dot-product between the $i$-th query and the $j$-th document token. The use of the $\max$ operator ensures that only the most relevant document tokens are considered for each query token, effectively filtering out those that are locally irrelevant.

Training is performed by jointly optimizing both the query and the document encoder with a symmetric InfoNCE loss~\cite{radford2021learning}, where global query-document cosine similarities are replaced with the score defined in Eq.~\ref{eq:fine-grained loss}. A key difference from the common approach of using a symmetric representation for both queries and documents, our encoders optimize two independent sets of learnable parameters, thus allowing for an asymmetric representation.

\section{Experiments}
\label{sec:experiments}

\tinytit{Training and Evaluation Datasets} We train and evaluate our models on the M2KR benchmark, which is a diverse collection of datasets repurposed for multimodal information retrieval, including MSMARCO~\cite{nguyen2016msmarco}, WIT~\cite{srinivasan2021wit}, IGLUE~\cite{bugliarello2022iglue}, KVQA~\cite{shah2019kvqa}, CC3M~\cite{sharma2018conceptual}, OVEN~\cite{hu2023open}, LLaVA~\cite{liu2024visual}, OKVQA~\cite{marino2019ok}, InfoSeek~\cite{chen2023can}, and Encyclopedic-VQA~\cite{mensink2023encyclopedic}.
Since we focus on a retrieval setting where both queries and documents are multimodal, we enrich the documents from Encyclopedic-VQA, InfoSeek, OVEN, and OKVQA with images (see supplementary material for details). For evaluation, we also employ the M-BEIR bechmark~\cite{wei2024uniir}, which comprises eight multimodal retrieval tasks and ten datasets from a variety of domains and image sources. To be consistent with our multimodal setup, we evaluate \ours on the InfoSeek and OVEN splits of M-BEIR$_\text{local}$, where the query is always multimodal, and though document images may be missing in some instances. 

\tit{Evaluation Metrics} 
Following the evaluation protocol defined in M2KR~\cite{lin2024preflmr}, we measure the performance of our models in terms of recall at $K$, where $K$ is adopted according to the experimental protocol of the specific sub-dataset. For visual question answering datasets, following~\cite{lin2024preflmr}, we also report the pseudo recall metric which counts a retrieved document as relevant whenever it contains the answer.

\tit{Implementation Details}
We train our models via contrastive learning with the Adam~\cite{kingma2015adam} optimizer, a $5\times10^{-5}$ learning rate with a cosine scheduler, and a total batch size of $512$. We leverage DisCo~\cite{chen2023disco} to optimize memory. Differently from previous works~\cite{lin2024preflmr}, we skip in-batch negative sampling and train in a single stage on the full data distribution.
If not specified, we employ frozen CLIP ViT-L14 visual and textual backbones.
A linear projection for each layer is applied to CLIP's textual and visual outputs before feeding it through the recurrent cell of \ours. 
\ours use a hidden size $d$ equal to $768$ and $1024$ for the base and larger Vision Transformer variants, respectively. The number of learnable input tokens $k$ is set to $32$. Following prior work~\cite{santhanam2022colbertv2, lin2023fine, lin2024preflmr}, fine-grained late-interaction tokens are projected to a hidden size of 
$128$, denoted as $\overline{d}$. 

\begin{table*}
\centering
\setlength{\tabcolsep}{.3em}
\resizebox{\linewidth}{!}{%
\begin{tabular}{lc c cc cc cc cc cc c cc cc c cc}
\toprule 
& & & \multicolumn{1}{c}{\textbf{WIT}} & & \multicolumn{1}{c}{\textbf{IGLUE}} & & \multicolumn{1}{c}{\textbf{KVQA}} & & \multicolumn{1}{c}{\textbf{OVEN}} & & \multicolumn{1}{c}{\textbf{LLaVA}} & & \multicolumn{2}{c}{\textbf{InfoSeek}} & & \multicolumn{2}{c}{\textbf{E-VQA}} & & \multicolumn{2}{c}{\textbf{OKVQA}} \\
\cmidrule{4-4} \cmidrule{6-6} \cmidrule{8-8} \cmidrule{10-10} \cmidrule{12-12} \cmidrule{14-15} \cmidrule{17-18} \cmidrule{20-21}
\textbf{Model} & \textbf{Visual Encoder} & & R@10 & & R@1 & & R@5 & & R@5 & & R@1 & & R@5 & PR@5 & & R@5 & PR@5 & & R@5 & PR@5 \\
\midrule
CLIP (Feature Averaging)~\cite{radford2021learning} & CLIP ViT-L & & 0.7 & & 1.6 & & 1.9 & & 41.1 & & 21.3 & & 33.1 & 47.9 & & 0.3 & 12.2 & & 5.8 & 48.0 \\
UniIR (Feature Fusion)~\cite{wei2024uniir} & BLIP ViT-L & & 18.6 & & 30.9 & & 13.5 & & 63.6 & & 47.8 & & 24.5 & 46.0 & & 17.0 & 35.5 & & 9.3 & 58.1 \\ 
\midrule
FLMR~\cite{lin2023fine} & CLIP ViT-B & & 23.8 & & - & & 31.9 & & 40.5 & & 56.4 & & - & 47.1 & & - & - & & - & \textbf{68.1} \\
Pre-FLMR~\cite{lin2024preflmr} & CLIP ViT-B & & 41.7 & & 57.3 & & 28.6 & & 46.3 & & 67.2 & & - & \textbf{48.8} & & - & \textbf{67.9} & & - & 66.1 \\
CLIP (Unimodal) & CLIP ViT-B & & 47.6 & & 59.1 & & \textbf{33.7} & & 54.2 & & 28.4 & & 15.8 & 35.4 & & 9.7 & 23.0 & & 2.5 & 39.9 \\
CLIP (Feature Fusion) & CLIP ViT-B & & 41.6 & & 56.6 & & 22.0 & & 59.8 & & 58.0 & & 19.3 & 40.4 & & 21.2 & 40.5 & & 9.6 & 56.0 \\
\rowcolor{ourcolor}
\textbf{\ours (Ours)} & CLIP ViT-B & & \textbf{60.1} & & \textbf{73.9} & & 26.9 & & \textbf{72.9} & & \textbf{76.6} & & \textbf{30.2} & 48.1 & & \textbf{33.0} & 48.9 & & \textbf{13.9} & 58.3 \\
\midrule
Pre-FLMR~\cite{lin2024preflmr} & CLIP ViT-L & &  60.5 & & 69.2 & & 43.6 & & 59.8 & & 71.8 & & - & 57.9 & & - & \textbf{70.8} & & - & \textbf{68.5} \\
CLIP (Unimodal) & CLIP ViT-L & & 67.9 & & 73.6 & & 56.1 & & 69.1 & & 44.8 & & 25.7 & 44.2 & & 17.2 & 29.8 & & 4.3 & 37.9 \\
CLIP (Feature Fusion) & CLIP ViT-L & & 68.2 & & 76.9 & & 47.5 & & 76.0 & & 63.6 & & 38.2 & 54.7 & & 35.6 & 52.6 & & 12.1 & 59.4 \\
\rowcolor{ourcolor}
\textbf{\ours (Ours)} & CLIP ViT-L & & \textbf{73.4} & & \textbf{81.8} & & \underline{\textbf{63.5}} & & \textbf{82.0} & & \underline{\textbf{79.9}} & & \textbf{47.0} & \textbf{60.5} & & \textbf{44.5} & 57.9 & & \underline{\textbf{20.2}} & 66.2 \\
\midrule
Pre-FLMR~\cite{lin2024preflmr} & OpenCLIP ViT-H & & 60.5 & & 71.2 & & 39.4 & & 61.5 & & 72.3 & & - & 59.5 & & - & \textbf{71.7} & & - & \textbf{68.1} \\
CLIP (Feature Fusion) & OpenCLIP ViT-H & & 67.3 & & 78.4 & & 53.8 & & 81.7 & & 65.1 & & \textbf{51.9} & \underline{\textbf{64.0}} & & 38.3 & 54.7 & & 11.2 & 59.4 \\
\rowcolor{ourcolor}
\textbf{\ours (Ours)} & OpenCLIP ViT-H & & \textbf{71.4} & & \textbf{80.0} & & \textbf{59.3} & & \textbf{83.0} & & \textbf{79.8} & & 47.3 & 60.7 & & \textbf{44.8} & 57.8 & & \textbf{18.2} & 63.4 \\
\midrule
Pre-FLMR~\cite{lin2024preflmr} & OpenCLIP ViT-G & & 61.5 & & 71.5 & & 42.1 & & 63.4 & & 72.4 & & - & 59.6 & & - & \underline{\textbf{73.1}} & & - & \underline{\textbf{68.6}} \\
CLIP (Feature Fusion) & OpenCLIP ViT-G & & \underline{\textbf{77.1}} & & 78.7 & & 59.0 & & 83.5 & & 67.6 & & 48.4 & 61.8 & & 43.8 & 56.5 & & 10.6 & 60.4 \\
\rowcolor{ourcolor}
\textbf{\ours (Ours)} & OpenCLIP ViT-G & & 75.1 & & \underline{\textbf{82.2}} & & \textbf{60.6} & & \underline{\textbf{84.0}} & & \textbf{79.2} & & \underline{\textbf{52.0}} & \textbf{62.5} & & \underline{\textbf{48.6}} & 60.2 & & \textbf{19.0} & 63.8 \\
\bottomrule
\end{tabular}
}
\vspace{-0.1cm}
\caption{Experimental results on the M2KR benchmark~\cite{lin2024preflmr}, comparing \ours to baselines and competitors when varying the visual backbone. Bold font denotes the best results with the same backbone, while underlined values indicate the highest overall scores.}
\label{tab:m2kr}
\vspace{-0.4cm}
\end{table*}

\subsection{Baselines and Competitors}
To ensure a comprehensive comparison, we introduce three baselines within our experimental setup.

\tit{CLIP (Feature Averaging)}
This approach leverages the zero-shot capabilities of CLIP, which we adapt for the multimodal query/document setting by averaging the unimodal feature outputs from each encoder. On the query side, the query text and image are processed separately by two distinct, frozen CLIP encoders, and the resulting pooled feature vectors are then combined by their average. This same approach is applied on the document side. 

\tit{CLIP (Unimodal)} This is a straightforward extension of CLIP to the fine-grained late-interaction paradigm. We use all the tokens from the last layer of the CLIP visual encoder to represent the query, whereas the document is encoded by taking the output of the last layer of the CLIP textual encoder. Because CLIP has not been trained using a fine-grained relevance score (cfr. Eq.~\ref{eq:fine-grained loss}), we apply LoRA~\cite{hu2021lora} to the textual and visual encoders and train them along with the late-interaction linear projection $\textbf{W}_{final}$.

\tit{CLIP (Feature Fusion)} Inspired by the feature fusion approach proposed in~\cite{wei2024uniir}, we build a multimodal retriever by training a cross-attention layer with residual connection on top of two frozen CLIP encoders. The attention queries in the cross-attention layers are the visual tokens on the query side and the textual tokens on the document side. For unimodal textual documents, we feed a black image to the CLIP visual encoder and mask out the cross-attention output before adding it to the residual connection.

\tit{FLMR~\cite{lin2023fine} and PreFLMR~\cite{lin2024preflmr}} These are multimodal retrievers pre-trained on a vision-language corpus of over ten million items where the query is multimodal, while the document side contains only textual passages. While FLMR only uses the \texttt{CLS} embedding from the visual backbone as the image representation, PreFLMR enhances this by also extracting embeddings of image patches from the penultimate layer to provide a more detailed visual representation. Notably, PreFLMR adopts a three-stage training strategy, with the second one dedicated exclusively to Encyclopedic-VQA to better handle its higher diversity, difficulty, and dataset size compared to other KB-VQA datasets.

\tit{UniIR~\cite{wei2024uniir}} They propose strategies for encoding multimodal queries and documents, by leveraging pre-trained models like CLIP~\cite{radford2021learning} and BLIP~\cite{li2022blip} to integrate different modalities, with the goal of creating a unified retriever for diverse tasks. Here we consider the feature-level fusion version, that utilizes only the features from the last layer while fine-tuning both visual and textual backbones.

\subsection{Comparison with the State of the Art}
\tinytit{Results on M2KR}
In Table~\ref{tab:m2kr}, we compare our method with the previously described baselines and with FMLR and Pre-FLMR, while also exploring the effect of scaling the visual backbone. As it can be seen, \ours achieves the best performance across all considered backbone sizes and on the great majority of considered datasets, consistently outperforming both competitors and baselines. 
For example, compared to PreFLMR~\cite{lin2024preflmr}, \ours demonstrates a significant performance boost on the OVEN dataset, where we improve of $+26.6$ and $+22.2$ R@5 points when using the CLIP ViT-B and ViT-L backbones. Similar improvements can be observed also across WIT, IGLUE, KVQA, LLaVA and InfoSeek. On E-VQA and OKVQA our proposed approach achieves better results than FLMR and Pre-FLMR in terms of R@5, while achieving comparable or slightly lower results in terms of PR@5. This is due to the training strategy employed by the two competitors, which privileges these two datasets during batch sampling. Overall, our results confirm the effectiveness of the feature fusion strategy and of the proposed Transformer-based recurrent cell.
Overall, scaling the visual backbone enhances retrieval performance. The only exception is observed with the OpenCLIP ViT-H backbone, which performs lower than the comparatively smaller CLIP ViT-L model. This observation aligns with findings in PreFLMR, suggesting that larger visual encoders may not universally enhance performance.

\tit{Results on the M-BEIR Benchmark}
We further test the generalization ability of our proposed approach on the OVEN and InfoSeek splits of M-BEIR~\cite{wei2024uniir}, where the queries is always multimodal, and the documents are either textual or paired text-image passages. 
\begin{table}[t]
\centering
\setlength{\tabcolsep}{.32em}
\resizebox{\linewidth}{!}{%
\begin{tabular}{lc cc c cc}
\toprule 
& &  \multicolumn{2}{c}{$\mathbf{(q^T, q^V) \rightarrow d^T}$} & & \multicolumn{2}{c}{$\mathbf{(q^T, q^V) \rightarrow (d^T, d^V)}$} \\
\cmidrule{3-4} \cmidrule{6-7} 
& & \textbf{OVEN} & \textbf{InfoSeek} & & \textbf{OVEN} & \textbf{InfoSeek} \\
\midrule
\rowcolor{LightGray}
\multicolumn{7}{l}{\textit{Zero-Shot}} \\
CLIP (ViT-L)~\cite{radford2021learning} & & 24.2 & 20.5 & & 38.8 & 26.4 \\
SigLIP (ViT-L)~\cite{zhai2023sigmoid} & & 29.7 & 25.2 & & 41.7 & 27.4 \\
BLIP-2~\cite{li2023blip} & & 12.2 & 5.5 & & 27.3 & 15.8 \\
\midrule
\rowcolor{LightGray}
\multicolumn{7}{l}{\textit{Trained on M2KR}} \\
Pre-FLMR~\cite{lin2024preflmr} & & 5.0 & 10.2 & & - & - \\
CLIP (Feature Fusion) & & 13.5 & 22.1 & & 31.1 & 18.6 \\
\rowcolor{ourcolor}
\textbf{\ours (Ours)} & & \textbf{26.8} & \textbf{28.3} & & \textbf{51.9} & \textbf{30.8} \\
\midrule
\rowcolor{LightGray}
\multicolumn{7}{l}{\textit{Trained or Fine-tuned on M-BEIR}} \\
UniIR (Feature Fusion)~\cite{wei2024uniir} & & \textbf{41.0} & 22.4 & & 55.8 & 33.0 \\
CLIP (Feature Fusion) & & 23.5 & 20.3 & & 49.5 & 30.7 \\
\rowcolor{ourcolor}
\textbf{\ours (Ours)} & & 35.6 & \textbf{25.6} & & \textbf{56.3} & \textbf{36.5} \\
\bottomrule
\end{tabular}
}
\vspace{-0.1cm}
\caption{Experimental results in terms of R@5 on the subsets of the M-BEIR benchmark~\cite{wei2024uniir} with multimodal queries.}
\label{tab:mbeir}
\vspace{-0.4cm}
\end{table}
In Table~\ref{tab:mbeir}, we again compare \ours against other feature fusion models (\ie, PreFLMR~\cite{lin2024preflmr}, UniIR~\cite{wei2024uniir}, and our CLIP feature fusion baseline). Notably, in this experiment, PreFLMR employs OpenCLIP ViT-G~\cite{cherti2023reproducible} as its visual backbone, while the others leverage CLIP ViT-L. For training and evaluation, we stick with the official instruction set of M-BEIR, which slightly differs from those of M2KR, and we concatenate it to the query text. 
Among methods trained on the M2KR benchmark, \ours demonstrates the strongest performance, showing substantial gains over PreFLMR. Notably, when trained on M2KR, our model outperforms UniIR on the InfoSeek split where no document images are provided, reaching R@5 $28.3$ compared to UniIR which achieves R@5 $22.4$. After fine-tuning of around half an epoch on the InfoSeek and OVEN splits of M-BEIR, \ours beats UniIR on both datasets, when including images on the document side. For instance, our model reaches an R@5 of $56.3$ and $36.5$ on OVEN and InfoSeek. This further highlights the importance of multiple modalities when embedding documents.
\begin{table*}[t]
\centering
\setlength{\tabcolsep}{.35em}
\resizebox{0.85\linewidth}{!}{%
\begin{tabular}{lc c cc cc cc cc cc c cc cc c cc}
\toprule 
& & \multicolumn{1}{c}{\textbf{WIT}} & & \multicolumn{1}{c}{\textbf{IGLUE}} & & \multicolumn{1}{c}{\textbf{KVQA}} & & \multicolumn{1}{c}{\textbf{OVEN}} & & \multicolumn{1}{c}{\textbf{LLaVA}} & & \multicolumn{2}{c}{\textbf{InfoSeek}} & & \multicolumn{2}{c}{\textbf{E-VQA}} & & \multicolumn{2}{c}{\textbf{OKVQA}} \\
\cmidrule{3-3} \cmidrule{5-5} \cmidrule{7-7} \cmidrule{9-9} \cmidrule{11-11} \cmidrule{13-14} \cmidrule{16-17} \cmidrule{19-20}
\textbf{Model} & & R@10 & & R@1 & & R@5 & & R@5 & & R@1 & & R@5 & PR@5 & & R@5 & PR@5 & & R@5 & PR@5 \\
\midrule
\rowcolor{LightGray}
\multicolumn{20}{l}{\textit{Loss Function}} \\
w/o fine-grained loss & & 41.9 & & 51.1 & & 6.3 & & 73.5 & & 24.7 & & 25.3 & 45.7 & & 2.4 & 13.5 & & 11.9 & 46.0 \\
\rowcolor{ourcolor}
\textbf{\ours (Ours)} && \textbf{73.4} & & \textbf{81.8} & & \textbf{63.5} & & \textbf{82.}0 & & \textbf{79.9} & & \textbf{47.0} & \textbf{60.5} & & \textbf{44.5} & \textbf{57.9} & & \textbf{20.2} & \textbf{66.2}  \\
\midrule
\rowcolor{LightGray}
\multicolumn{20}{l}{\textit{Effect of Multimodal Document Encodings}} \\
w/o candidate images & & 73.1 & & \textbf{81.8} & & 61.4 & & 77.5 & & \textbf{80.0} & & 46.8 & 59.2 & & 37.0 & 51.9 & & \textbf{24.1} & \textbf{68.6}\\
\rowcolor{ourcolor}
\textbf{\ours (Ours)} && \textbf{73.4} & & \textbf{81.8} & & \textbf{63.5} & & \textbf{82.0} & & 79.9 & & \textbf{47.0} & \textbf{60.5} & & \textbf{44.5} & \textbf{57.9} & & 20.2 & 66.2  \\
\midrule
\rowcolor{LightGray}
\multicolumn{20}{l}{\textit{Effect of Recurrence}} \\
w/o recurrence & & 69.8 & & 81.2 & & 62.4 & & 80.6 & & 74.5 & & 40.5 & 56.8 & & 42.7 & 56.2 & & 16.4 & 61.8 \\
w/ recurrence in first 4 layers  & & 42.1 & & 56.2 & & 22.2 & & 54.2 & & 66.2 & & 10.1 & 34.3 & & 19.0 & 38.0 & & 12.8 & 51.7 \\
w/ recurrence in last 4 layers & & 73.0 & & \textbf{82.0} & & 63.4 & & \textbf{82.3} & & 79.7 & & 45.7 & 59.2 & & 43.2 & 56.7 & & 19.6 & \textbf{66.4} \\
\rowcolor{ourcolor}
\textbf{\ours (Ours)} && \textbf{73.4} & & 81.8 & & \textbf{63.5} & & 82.0 & & \textbf{79.9} & & \textbf{47.0} & \textbf{60.5} & & \textbf{44.5} & \textbf{57.9} & & \textbf{20.2} & 66.2  \\
\bottomrule
\end{tabular}
}
\vspace{-0.15cm}
\caption{Ablation study results. All experiments are with CLIP ViT-L for both visual and textual encoders.}
\label{tab:ablation_large}
\vspace{-0.35cm}
\end{table*}

\subsection{Ablation Studies and Analyses} 

\tinytit{Loss Function Analysis}
\ours is trained to optimize a fine-grained contrastive loss (Eq.~\ref{eq:final}).
For comparison, we train an identical model but using the standard contrastive loss, where relevance is computed as a single cosine similarity score. Specifically, we average the $k$ late-interaction tokens from the final recurrent iteration obtaining a single global feature vector.
As shown in Table~\ref{tab:ablation_large} (top), removing the fine-grained loss leads to a noticeable performance drop across the M2KR benchmark, highlighting the value of fine-grained interactions in enhancing retrieval accuracy.

\tit{Effect of Multimodal Encodings}
Given that some competitors do not incorporate document images, we conduct an ablation study to evaluate their impact on \ours. We train and evaluate \ours on M2KR without document images by replacing them in the OVEN, InfoSeek, Encyclopedic-VQA, and OKVQA datasets with black pictures, as a standard practice in the literature~\cite{wei2024uniir}. Consequently, we mask out their visual contributions while computing the next candidate state of the recurrent cell.
As shown in Table~\ref{tab:ablation_large} (middle), we experience substantial gains by enriching document representations with visual features. Notably, \ours performs better also on WIT and KVQA, which, by design, do not include images on the document side. Despite this, these datasets still benefit from the multimodal document training conducted across the entire benchmark, demonstrating the robustness and generalizability of our approach. The only significant drop is on the OKVQA dataset, likely due to the way document images are collected and the reduced number of samples from OKVQA in the training set, compared to OVEN and InfoSeek. 

\begin{figure*}[t]
    \centering
    \includegraphics[width=0.94\linewidth]{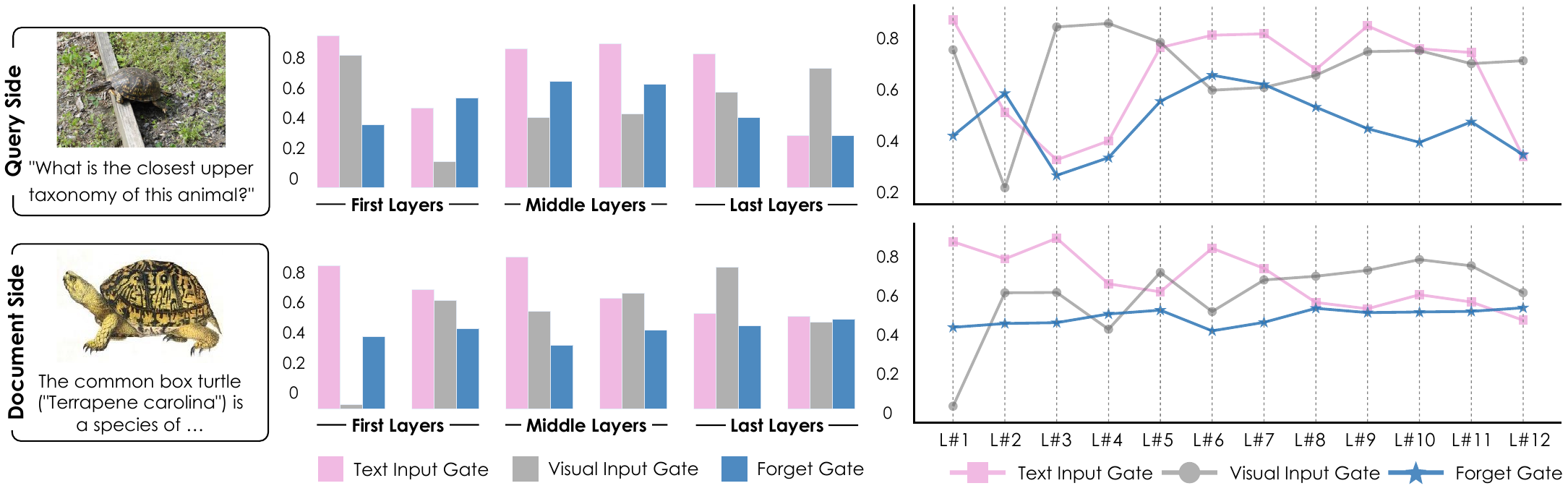}
    \vspace{-0.25cm}
    \caption{Qualitative analysis of gate activations. The left-side displays gate behavior during the encoding of a single query-document pair, while the right shows average activations over 2k examples from the M2KR InfoSeek test split.}
    \label{fig:gate_activations}
    \vspace{-0.4cm}
\end{figure*}

\tit{Effect of Recurrence}
We now discuss whether the intuition behind merging features from multiple layers through recurrence is well-founded. To this end, we train and evaluate \ours employing the proposed recurrent Transformer cell only once, applied to the final layer outputs from the textual and visual backbones. This setup allows us to isolate the impact of a single-layer recurrence versus our full multi-layer approach.
As shown in Table~\ref{tab:ablation_large} (bottom), removing multi-layer recurrence leads to a significant performance drop across all benchmarks, emphasizing its crucial role in retrieval performance.
Having established the value of recurrence, we question its application on all layers of the pre-trained backbones. Unsurprisingly, applying the recurrent cell on the last four layers of the backbones yields significantly better performance compared to applying it to the first four layers. 
This aligns with the intuition that deeper layers retain lower-level information while shallower layers struggle to capture high-level semantics
Moreover, while most queries in M2KR likely only require a global understanding of the query and image, leveraging information from all layers (\ie, \ours) emerges as the best strategy, particularly for challenging VQA datasets. These results confirm that integrating low-level features enhances the results of multimodal retrievers based on feature fusion.

\begin{table}[t]
\vspace{-0.1cm}
\centering
\setlength{\tabcolsep}{.15em}
\resizebox{\linewidth}{!}{%
\begin{tabular}{l cccc c ccc cc}
\toprule 
 & \multicolumn{4}{c}{\textbf{Training Info}} & & \multicolumn{3}{c}{\textbf{Inference Time (ms)}}\\
\cmidrule(lr){2-5} \cmidrule(lr){7-9}
\textbf{Model} & \multicolumn{2}{c}{Backbones} & \#GPUs & Hrs & & Forward & Retrieval & All $\downarrow$ & & \#QTokens \\
\midrule
Pre-FLMR~\cite{lin2024preflmr} & T \faFire & V \SnowflakeChevron & 4 & 864 & & 32.7 & 406.1 & 438.8 & & 320 \\
\midrule
CLIP (Unimodal) & T \faFire & V \faFire & 8 & 96 & & 16.9 & 17.2 & 34.1 & & 257 \\
CLIP (Feat. Fusion) & T \SnowflakeChevron & V \SnowflakeChevron & 4 & 80 & & 21.4 & 16.1 & 37.5 & & 257 \\
\rowcolor{ourcolor}
\textbf{\ours (Ours)} & T \SnowflakeChevron & V \SnowflakeChevron & 4 & 80 & & 31.4 & 3.5 & 34.9 & & 32 \\
\bottomrule
\end{tabular}
}
\vspace{-.15cm}
\caption{Computational analysis.}
\label{tab:training_inference}
\vspace{-0.4cm}
\end{table}

\tit{Computational analysis}
In Table~\ref{tab:training_inference}, we provide a computational analysis of \ours, comparing it against various baselines and a key competitor. 
All baselines are trained with the same data and ViT-L/14 backbone, except for Pre-FLMR, which uses ColBERTv2 for the text encoder. In terms of computational resources, \ours requires equal or fewer GPU hours compared to competitors and baselines that fine-tune one or both encoder backbones. During inference, the computational cost for a single query-document comparison varies across approaches. In fine-grained late interaction models, the number of dot product operations scales with the product of query and document tokens. However, for \ours, this is fixed at just $32$. In contrast, the concatenation of visual and textual tokens of PreFLMR leads to much larger query token counts, often exceeding $300$, while document tokens are capped at $512$. As a result, while \ours incurs a slight increase in encoding time due to its recurrent module, it compensates with superior efficacy and a faster retrieval phase, thanks to the reduced token count. Notably, the CLIP (Unimodal) baseline ignores images when encoding queries, explaining its faster forward.

\tit{Qualitative Analysis of Gate Activations}
Fig.~\ref{fig:gate_activations} presents the activation patterns of our learnable sigmoid gates within the recurrent cell of \ours, when processing various textual and visual layer representations. On the left, we show gate behavior for a single query-document encoding, while on the right, we show average activations over 2k samples from the InfoSeek split of M2KR. Higher forget gate values indicate information preservation from previous layers, while higher values in the unimodal input gates enable increased information flow from the respective modality. 
When encoding queries, gate activations fluctuate across layers, indicating that the model leverages all representations offered by the pre-trained backbones. Notably, forget gate values never drop to zero, ensuring continuous information retention. 
In contrast, during multimodal document encoding, gate activations are more stable and balanced. Early layers prioritize text, while deeper layers gradually increase forget gate values, enabling selective integration of relevant visual features. 
This behavior likely reflects a strategy to minimize visual noise in document representation, as images may not consistently align with the associated text.

\begin{table}[t]
\centering
\setlength{\tabcolsep}{.25em}
\resizebox{\linewidth}{!}{
\begin{tabular}{lcc cc c c cc c}
\toprule
& & & \multicolumn{3}{c}{\textbf{InfoSeek (top-1)}} & & \multicolumn{3}{c}{\textbf{InfoSeek (top-3)}} \\
\cmidrule{4-6} \cmidrule{8-10}
\textbf{LLM} & \textbf{Retriever} & & Un-Q & Un-E & All & & Un-Q & Un-E & All \\
\midrule
LLaMA-3-8B & - & & 0.5 & 0.7 & 0.6 & & 0.5 & 0.7 & 0.6 \\
LLaMA-3-8B & CLIP & & 14.7 & 14.6 & 14.6 & & 17.5 & 17.3 & 17.4 \\
LLaMA-3-8B & Pre-FLMR & & 15.5 & 13.5 & 14.5 & & 18.2 & 15.9 & 16.9 \\
\rowcolor{ourcolor}
LLaMA-3-8B & \textbf{\ours (Ours)} & & \textbf{21.0} & \textbf{15.5} & \textbf{17.9} & & \textbf{25.6} & \textbf{18.6} & \textbf{21.6} \\
\midrule
LLaMA-3.1-8B & - & & 0.8 & 0.6 & 0.7 & & 0.8 & 0.6 & 0.7 \\
LLaMA-3.1-8B & CLIP & & 14.5 & 14.7 & 14.6 & & 17.5 & 17.8 & 17.6 \\
LLaMA-3.1-8B & Pre-FLMR & & 16.5 & 14.5 & 15.5 & & 18.5 & 16.5 & 17.5 \\
\rowcolor{ourcolor}
LLaMA-3.1-8B & \textbf{\ours (Ours)} & & \textbf{21.0} & \textbf{15.8} & \textbf{18.1} & & \textbf{26.1} & \textbf{19.1} & \textbf{22.1} \\
\bottomrule
\end{tabular}
}
\vspace{-.15cm}
\caption{RAG results on InfoSeek with different retrievers.
}
\label{tab:vqa}
\vspace{-0.4cm}
\end{table}

\subsection{Retrieval-Augmented VQA}
Finally, we test the quality of the passages retrieved by our model on the challenging InfoSeek dataset~\cite{chen2023can} for knowledge-based visual question answering, which contains image-question pairs that require external knowledge retrieval~\cite{caffagni2024wiki,yan2024echosight}. Specifically, we employ the official validation split of the dataset and a knowledge base of 525k multimodal passages. To generate answers, we exploit two state-of-the-art text-only LLMs (\ie, LLaMA-3 and LLaMA-3.1~\cite{dubey2024llama}, both in their 8B version). 
As these models cannot directly process images, they must rely exclusively on their pre-trained knowledge or external retrieved passages to answer questions. 
In datasets like InfoSeek, the absence of retrieval significantly degrades performance, with scores nearly dropping to zero, as shown in Table~\ref{tab:vqa}. To enhance the generation capabilities of the LLMs, we provide the top-1 and top-3 retrieved passages using \ours, PreFLMR, or the original CLIP model.
As it can be seen, a standard cross-modal image-to-text retriever (\ie, CLIP) performs similarly to a multimodal feature fusion method such as PreFLMR, probably due to the reduced generalization capabilities of PreFLMR on entities unseen during training. 
By leveraging multimodal features from both queries and documents, \ours outperforms other retrievers, achieving the best results in both settings, highlighting the effectiveness of our approach in challenging retrieval tasks.
\section{Conclusion}
\label{sec:conclusion}
We introduced \ours, a novel multimodal retrieval method that fuses text and image features in both queries and documents. Unlike existing methods, \ours fuses representations from multiple layers of pre-trained backbones through a transformer-based recurrent cell. The output is a set of late-interaction tokens that are used to compute fine-grained query-document relevance scores. \ours showcases state-of-the-art performance on the challenging M2KR multimodal retrieval benchmark, outscoring other feature fusion methods on most datasets. We believe our approach of leveraging multi-layer features offers a promising direction for advancing neural information retrieval.

\vspace{0.15cm}
\tit{Acknowledgments}
\small{We acknowledge the CINECA award under the ISCRA initiative, for the availability of high-performance computing resources. This work has been conducted under a research grant co-funded by Leonardo S.p.A. and supported by the PNRRM4C2 project ``FAIR'', by the PRIN 2022-PNRR project ``MUCES'' (CUP E53D23016290001), and by the PNRR project ``ITSERR'' (CUP B53C22001770006), all funded by the EU - NextGenerationEU. The authors would also like to thank Federico Cocchi and Nicholas Moratelli for their valuable support.}

{
    \small
    \bibliographystyle{ieeenat_fullname}
    \bibliography{bibliography}
}

\newpage
\newpage
\normalsize
\appendix

\noindent In the following, we present further experiments and details about the proposed multimodal retriever \ours. 

\section{Additional Details on Experimental Setup}

\subsection{Adding Images to Documents}
In the original M2KR benchmark~\cite{lin2024preflmr}, while queries are always text-image pairs, documents entail only textual data (\ie, $\ \doc = d^T$). To enable the training of retrieval models capable of understanding multimodal documents, we enrich the splits of OVEN~\cite{hu2023open}, InfoSeek~\cite{chen2023can}, E-VQA~\cite{mensink2023encyclopedic}, and OKVQA~\cite{marino2019ok} by adding images in their respective documents (\ie, $\ \doc = (d^T, d^V)$). We only include images from the official collections found in literature, using the same knowledge base proposed along the original dataset when possible. These enriched documents, along with our models, will be publicly released. The number of document images collected for each dataset is detailed in Table~\ref{tab:doc_img_stats}.

\tit{OVEN} The dataset features a knowledge base of 6 million Wikipedia pages (Wiki6M), each uniquely identified by a \texttt{WikidataID} and potentially accompanied by an image. Text passages in M2KR are assigned a unique identifier in the format \texttt{Wiki6M\_WikidataID}, allowing us to use the \texttt{WikidataID} to associate the corresponding page in Wiki6M. If the page includes an image, it is selected as $d^V$ in the multimodal document. 

\tit{InfoSeek} The dataset is derived from OVEN and, as such, shares the same knowledge base. Each M2KR document begins with a \texttt{WikiTitle}, followed by the textual content. We use the \texttt{WikiTitle} as a query to match the corresponding page in Wiki6M and select its image as $d^V$.

\tit{E-VQA} The official knowledge base of E-VQA comprehends 2M Wikipedia articles. Similarly to InfoSeek, for each document, whose ID is in the form \texttt{WikiWeb\_WikiTitle\_WikiSectionID}, we search for the page with the same \texttt{WikiTitle} in the 2M collection. However, the knowledge base of E-VQA divides a page into multiple sections, and each section may have a dedicated image. Consequently, given a matching page, we look for the image related to the section identified by \texttt{WikiSectionID}. If that image is missing, we consider the first available image on the same page as $d^V$.

\tit{OKVQA} The documents of OKVQA in M2KR are extracted from Wikipedia articles as well, according to~\cite{lin2023fine}. The content of each document is preceded by the title of the article, as in InfoSeek. Given that no other source of images is available, we search into Wiki6M for Wikipedia pages whose title matches the \texttt{WikiTitle} found on each document. For a matching page, we attach its image to the corresponding OKVQA document. We highlight a possible drawback concerning document images in OKVQA.  In particular, we find that $15$\% and $4$\% of them are also found on the documents of OVEN and InfoSeek, respectively. Given that the train split of OKVQA is much smaller compared to those two datasets (see Table~\ref{tab:doc_img_stats}), it may happen that, during training, those images are more easily associated with textual documents from OVEN and InfoSeek, ignoring OKVQA. A potential confirmation of this hypothesis is that OKVQA is the only dataset where we report superior performance when training without images on the document side (see Table~\ref{tab:ablation_large}, middle).

\begin{table}[t]
\centering
\setlength{\tabcolsep}{.25em}
\resizebox{\linewidth}{!}{
\begin{tabular}{lcc cc cc cc cc}
\toprule
& & \multicolumn{2}{c}{\textbf{Train}} & & \multicolumn{2}{c}{\textbf{Val}} & & \multicolumn{2}{c}{\textbf{Test}} \\
\cmidrule{3-4} \cmidrule{6-7} \cmidrule{9-10}
\textbf{Dataset} & & \textit{w/ Images} & \textbf{Total} & &\textit{w/ Images} & \textbf{Total} & & \textit{w/ Images} & \textbf{Total} \\
\midrule
OVEN  & & 325,746 & 339,137 & & 113,723 & 119,136 & & ~~~3,067 & ~~~3,192 \\
InfoSeek & & 647,308 & 676,441 & & - & - & & 79,914 & 98,276 \\
E-VQA & & 164,518 & 167,369 & & ~~~9,722 & ~~~9,852 & & ~~~3,660 & ~~~3,750 \\
OKVQA & & ~~~4,253 & ~~~9,009 & & ~~~2,415 & ~~~5,046 & & ~~~2,415 & ~~~5,046\\
\bottomrule
\end{tabular}
}
\vspace{-.15cm}
\caption{Number of document with images and total number of documents in each split of OVEN, InfoSeek, E-VQA, and OKVQA.}
\label{tab:doc_img_stats}
\vspace{-0.15cm}
\end{table}

\begin{table}[t]
\centering
\setlength{\tabcolsep}{.2em}
\resizebox{\linewidth}{!}{
\begin{tabular}{lcc ccc ccc}
\toprule
& & \multicolumn{3}{c}{\textbf{Text Encoder}} & & \multicolumn{3}{c}{\textbf{Visual Encoder}} \\
\cmidrule{3-5} \cmidrule{7-9}
\textbf{Backbone} & & $\overline{L}$ & $L$ & \textbf{Layer Indices} & & $\overline{L}$ & $L$ & \textbf{Layer Indices} \\
\midrule
CLIP ViT-B & & 12 & 12 & all & & 12 & 12 & all \\

\multirow{2}{*}{CLIP ViT-L} & & \multirow{2}{*}{12} & \multirow{2}{*}{12} & \multirow{2}{*}{all} & & \multirow{2}{*}{24} & \multirow{2}{*}{12} & 0, 2, 4, 6, 8, 10, \\
& & & & & & & & 12, 14, 16, 18, 20, 22 \\

\multirow{2}{*}{OpenCLIP ViT-H} & & \multirow{2}{*}{24} & \multirow{2}{*}{12} & 1, 3, 5, 7, 9, 11, & & \multirow{2}{*}{32} & \multirow{2}{*}{12} & 0, 2, 5, 8, 11, 14, \\
& & & & 13, 15, 17, 19, 21, 23 & & & & 17, 20, 23, 26, 29, 31 \\
\multirow{2}{*}{OpenCLIP ViT-G} & & \multirow{2}{*}{32} & \multirow{2}{*}{16} & 0, 2, 4, 6, 8, 10, 12, 14, & & \multirow{2}{*}{48} & \multirow{2}{*}{16} & 2, 5, 8, 11, 14, 17, 20, 23, \\
& & & & 16, 18, 20, 22, 24, 26, 28, 31 & & & & 26, 29, 32, 35, 38, 41, 44, 47 \\
\bottomrule
\end{tabular}
}
\vspace{-.15cm}
\caption{Indices of the layers selected for each backbone encoder to be used as input to the recurrent cell. We denote the original depth of a backbone, measured in number of layers, as $\overline{L}$, and the selected depth for a given configuration of \ours as $L$.}
\label{tab:n_layers}
\vspace{-0.4cm}
\end{table}

\subsection{Architectural Details}
\tinytit{Handling Different Depths of the Backbones} According to the notation defined in Sec.~\ref{sec:method_ret}, in \ours we assume that the textual and visual backbones have the same number of layers $L$. However, in practice, this condition is rarely satisfied. For instance, our typical configuration is based on CLIP ViT-L, which features $12$ layers in the text encoder and $24$ in the visual encoder. To address this discrepancy, we opt for selecting an equal number of layers from each unimodal encoder. Whenever possible, we sample the indices of the selected layers with a uniform interval. For CLIP ViT-L, this approach involves selecting one layer every two from the visual encoder. If a uniform interval is unfeasible, we manually adjust the number of layers. For OpenCLIP ViT-H and OpenCLIP ViT-G, we include $12$ and $16$ layers respectively from each backbone. Please refer to Table~\ref{tab:n_layers} for the exact layer indices selected for each configuration of \ours. With these premises, we highlight that the second and third ablation experiments concerning the \textit{Effect of Recurrence} in Table~\ref{tab:ablation_large} (bottom) assume that the number of layers for each backbone is already adjusted to the same depth $L$. That is, when we apply recurrence on the first four layers, it means that the recurrent cell operates on the first, third, fifth, and seventh layers of the visual backbone (\ie, the layer indices are 0, 2, 4, and 6).

\tinytit{Training Data} Our models are trained on the full M2KR dataset, with the exception of MSMARCO, as it does not include images on the query side. Following PreFLMR~\cite{lin2024preflmr}, we adjust the sampling proportions for each dataset. First, because InfoSeek contains many questions related to the same Wikipedia entity, we downsample the training split to obtain a roughly equal number of questions for each entity (around 76k samples). Second, we upsample E-VQA and OKVQA: 
duplicating E-VQA samples and repeating OKVQA samples nine times per epoch. Notably, duplicating E-VQA does not negatively affect performance on other datasets, nor does it lead to overfitting on its test set. For OKVQA, the upsampling is justified by the limited size of its training split, and we apply the same upsampling factor used in the third training stage of PreFLMR.

\tinytit{Additional Hyperparameters} The learnable gates of \ours can be adjusted by setting the forget and input gate bias, \ie, \ $b_f$ and $b_i$ respectively (cf. Eq.~\ref{eq:forget_gate}). A common choice is to choose $b_f$ equal to $1$ and $b_i$ equal to $-1$, which has the effect of biasing the model toward remembering the past. However, we empirically found no benefit in such a choice, as \ours natively learns to keep information from its past, that is, to retain information from shallower layers. So, we set both the forget and input gate bias to zero.
Following prior work~\cite{hutchins2022block}, the weights of the learnable matrices $\pmb{W}^T_F$, $\pmb{W}^V_f$, $\pmb{W}^V_i$, and $\pmb{W}^V_i$ in Eq.~\ref{eq:forget_gate}, that modulate the gates, are initially drawn from a truncated normal distribution. 

\tinytit{Efficient Indexing for Inference} 
For evaluating our models, we index the multimodal document collections of the target benchmark using PLAID~\cite{santhanam2022plaid}, which supports the fine-grained late-interaction paradigm. We refer readers to the original paper for detailed information about the indexing process. The only exception is when using CLIP for indexing the knowledge base for retrieval-augmented VQA in Table~\ref{tab:vqa}. Since CLIP generates a single embedding for both queries and documents, we build a Faiss index~\cite{johnson2019billion} to enable efficient dot-product similarity searches.

\subsection{Retrieval-Augmented VQA}
In the paper, we evaluate \ours for retrieval-augmented VQA on InfoSeek, which requires answering visual questions related to Wikipedia entities. The dataset has two subsets:
unseen question and unseen entity, which entail either questions or entities not found in the training set. 
Answers are found on the entity’s Wikipedia page within the 6M knowledge base~\cite{hu2023open} used by InfoSeek and OVEN in M2KR. Given the difficulty of the task, retrieval is essential.

\begin{table}[t]
\centering
\setlength{\tabcolsep}{.25em}
\resizebox{\linewidth}{!}{
\begin{tabular}{lcc cc c c cc c}
\toprule
& & & \multicolumn{3}{c}{\textbf{InfoSeek (top-1)}} & & \multicolumn{3}{c}{\textbf{InfoSeek (top-3)}} \\
\cmidrule{4-6} \cmidrule{8-10}
\textbf{MLLM} & \textbf{Retriever} & & Un-Q & Un-E & All & & Un-Q & Un-E & All \\
\midrule
LLaVA-v1.5 & - & & 6.9 & 7.3 & 7.1 & & 6.9 & 7.3 & 7.1  \\
LLaVA-v1.5 & CLIP & & 18.6 & 17.6 & 18.1 & & 21.0 & 20.1 & 20.6 \\
LLaVA-v1.5 & PreFLMR & & 17.4 & 15.8 & 16.6 & & 19.3 & 17.4 & 18.3 \\
\rowcolor{ourcolor}
LLaVA-v1.5 & \textbf{\ours (Ours)} & & \textbf{24.1} & \textbf{18.1} & \textbf{20.7} & & \textbf{28.1} & \textbf{21.1} & \textbf{24.1} \\
\midrule
LLaVA-MORE & - & & 7.3 & 7.4 & 7.4 & & 7.3 & 7.4 & 7.4 \\
LLaVA-MORE & CLIP & & 16.9 & 16.1 & 16.5 & & 19.9 & 18.7 & 19.3 \\
LLaVA-MORE & PreFLMR & & 17.1 & 15.4 & 16.2 & & 19.2 & 17.2 & 18.1 \\
\rowcolor{ourcolor}
LLaVA-MORE & \textbf{\ours (Ours)} & & \textbf{23.8} & \textbf{16.8} & \textbf{19.7} & & \textbf{28.5} & \textbf{20.3} & \textbf{23.8} \\
\bottomrule
\end{tabular}
}
\vspace{-.15cm}
\caption{Retrieval-augmented generation results on InfoSeek when using different information retrievers.}
\label{tab:vqa_mllm}
\vspace{-0.35cm}
\end{table}

\tit{Controlled Knowledge Base} To build our experimental knowledge base, we select from the original 6M collection all the Wikipedia pages whose entity is referenced in a question from the validation split. To expand it, we randomly sample additional entities, totalling 50K Wikipedia entities. Each Wikipedia page is chunked into shorter text passages of approximately 100 words~\cite{karpukhin2020dense}. Following M2KR, we prepend the Wikipedia title of the corresponding page to each passage, ending up with the format: 
\texttt{title: WikiTitle content: [...]}. 
\noindent
If the page includes an image, we attach it to every passage obtained from the same page, so that it can be used by \ours as the document image $d^V$. Our knowledge base includes 525,177 passages and will be open-sourced to support future research.

\tit{Retrieval Settings} We index our controlled knowledge base with either CLIP, PreFLMR, or \ours. Note that \ours is the only model that can leverage document images. On the query side, we prepend the textual query of a visual question from InfoSeek with a randomly chosen instruction, according to the M2KR format \footnote{For instance: 

\noindent\texttt{Using the provided image, obtain documents that address the subsequent question: \{question\}}.}. When embedding queries with CLIP, we ignore the question and only process the image with the visual encoder. 
To generate answers, we include the top-$k$ retrieved textual passages in the prompt of the chosen answer generator (\eg, a pre-trained LLM such as LLaMA-3). We experiment with $k$ equal to either 1 or 3. The prompt used is the following:

\vspace{0.04cm}
\resizebox{0.96\linewidth}{!}{
\begin{minipage}{\linewidth}
\noindent\texttt{Answer the question given the following context.}

\noindent\texttt{Context:}

\noindent \texttt{\{C1\}}

\noindent \texttt{\#\#}

\noindent \texttt{\{...\}}

\noindent \texttt{\#\#}

\noindent \texttt{\{C$k$\}}

\noindent \texttt{\#\#}

\noindent\texttt{Question: \{question\}}

\noindent\texttt{If the context does not help with the question, try to shortly answer it anyway.}

\noindent\texttt{Do not answer anything else.}

\noindent\texttt{Short answer:}
\end{minipage}
}
\vspace{0.2cm}

\noindent We omit the \textit{system} preamble because it is specific to the given LLM. Additionally, we augment the prompt with 3-shot examples, one for each question type that can be found in InfoSeek~\cite{chen2023can}: string, numerical, and date. The examples are randomly drawn from the training set.

\tit{Experiments with Multimodal LLMs} Table~\ref{tab:vqa_mllm} shows additional retrieval-augmented VQA results on InfoSeek. Differently from Table~\ref{tab:vqa}, here we use a Multimodal LLM (MLLM)~\cite{caffagni2024revolution} that can \textit{see} the query image as the answer generator. In particular, we employ the popular LLaVA-v1.5~\cite{liu2024visual}, based on Vicuna-7B, and the more recent LLaVA-MORE\footnote{\url{https://github.com/aimagelab/LLaVA-MORE}}, which is based on LLaMA-3.1-8B~\cite{dubey2024llama}. In general, MLLMs achieve better performance than LLMs, even though also in this context removing retrieval is severely detrimental. \ours confirms itself as the best information retriever, outscoring PreFLMR by as many as $5.8$ points in the top-3 settings when paired with LLaVA-v1.5. 

\begin{table*}[t]
\centering
\setlength{\tabcolsep}{.3em}
\resizebox{0.9\linewidth}{!}{
\begin{tabular}{lc c cc cc cc cc cc c cc cc c cc}
\toprule 
& & \multicolumn{1}{c}{\textbf{WIT}} & & \multicolumn{1}{c}{\textbf{IGLUE}} & & \multicolumn{1}{c}{\textbf{KVQA}} & & \multicolumn{1}{c}{\textbf{OVEN}} & & \multicolumn{1}{c}{\textbf{LLaVA}} & & \multicolumn{2}{c}{\textbf{InfoSeek}} & & \multicolumn{2}{c}{\textbf{E-VQA}} & & \multicolumn{2}{c}{\textbf{OKVQA}} \\
\cmidrule{3-3} \cmidrule{5-5} \cmidrule{7-7} \cmidrule{9-9} \cmidrule{11-11} \cmidrule{13-14} \cmidrule{16-17} \cmidrule{19-20}
\textbf{Model} & & R@10 & & R@1 & & R@5 & & R@5 & & R@1 & & R@5 & PR@5 & & R@5 & PR@5 & & R@5 & PR@5 \\
\midrule
\rowcolor{LightGray}
\multicolumn{20}{l}{\textit{Comparison with Naive Multi-Layer Feature Fusion}} \\
CLIP (Multi-Layer Feature Fusion) & & 59.9 & & 72.1 & & 38.4 & & 65.8 & & 59.9 & & 30.2 & 48.6 & & 21.4 & 42.2 & & 10.4 & 59.5 \\
\rowcolor{ourcolor}
\textbf{\ours (Ours)} & & \textbf{73.4} & & \textbf{81.8} & & \textbf{63.5} & & \textbf{82.0} & & \textbf{79.9} & & \textbf{47.0} & \textbf{60.5} & & \textbf{44.5} & \textbf{57.9} & & \textbf{20.2} & \textbf{66.2}  \\
\midrule
\rowcolor{LightGray}
\multicolumn{20}{l}{\textit{Effect of Recurrence on Query and Document Sides}} \\
w/o recurrence (query side) & & \textbf{75.4} & & 81.6 & & 58.8 & & 79.6 & & 75.9 & & 33.3 & 52.1 & & 35.1 & 50.9 & & 15.4 & 60.3 \\
w/o recurrence (document side) & & 74.0 & & 79.4 & & 62.7 & & 80.2 & & 76.8 & & \textbf{50.0} & \textbf{61.9} & & 35.5 & 51.6 & & 14.9 & 63.9 \\
\rowcolor{ourcolor}
\textbf{\ours (Ours)} && 73.4 & & \textbf{81.8} & & \textbf{63.5} & & \textbf{82.0} & & \textbf{79.9} & & 47.0 & 60.5 & & \textbf{44.5} & \textbf{57.9} & & \textbf{20.2} & \textbf{66.2}  \\
\midrule
\rowcolor{LightGray}
\multicolumn{20}{l}{\textit{Effect of Text Encoder}} \\
w/ ColBERTv2 & & \textbf{73.9} & & 79.3 & & 48.6 & & 79.6 & & 79.6 & & 40.0 & 58.9 & & 43.4 & \textbf{59.0} & & 19.0 & 64.1 \\
\rowcolor{ourcolor}
\textbf{\ours (Ours)} & & 73.4 & & \textbf{81.8} & & \textbf{63.5} & & \textbf{82.0} & & \textbf{79.9} & & \textbf{47.0} & \textbf{60.5} & & \textbf{44.5} & 57.9 & & \textbf{20.2} & \textbf{66.2}  \\
\bottomrule
\end{tabular}
}
\vspace{-0.15cm}
\caption{Additional ablation study results. All experiments are with CLIP ViT-L for both visual and textual encoders.}
\label{tab:ablation_large_supp}
\vspace{-0.3cm}
\end{table*}

\section{Additional Experimental Results} 
\tinytit{Comparison with Naive Multi-Layer Feature Fusion}
The intuition of combining features from multiple layers is, in principle, independent from the proposed recurrent cell architecture behind \ours. We thus propose a simpler baseline, that first collects the features from each layer of the CLIP textual and visual encoder. Then, the features from different layers are averaged and transformed with a learnable linear projection, obtaining a sequence of textual and visual tokens. These final representations are fused together via cross-attention, resulting in a multimodal sequence of tokens for fine-grained late-interaction retrieval. We report the results of this experiment in Table~\ref{tab:ablation_large_supp} (top). According to that, it can be concluded that fusing features from multiple layers by simply taking the mean is not competitive compared to the recurrent approach proposed by \ours.

\tit{Effect of Recurrence on Query and Document Sides}
From a qualitative analysis of the forget and input gate activations of \ours (see Fig.~\ref{fig:gate_activations}), we observe that the learned gates behave more uniformly across different layers when encoding multimodal documents. On the contrary, when processing multimodal queries, the gates experience larger fluctuations. We thus raise the question of whether the flexibility of the proposed multi-layer feature fusion strategy is necessary also to encode multimodal documents. To address this point, we train a variant of \ours, where we replace the recurrent cell in the document encoder with a cross-attention layer that is applied only to the features from the last layer of the pre-trained backbones. This is equivalent to adopting the proposed architecture of \ours to encode queries, and the CLIP (Feature Fusion) architecture proposed in Sec.~\ref{sec:experiments} as a baseline to process documents. For completeness, we also try the opposite, \ie, training CLIP (Feature Fusion) on the query side and \ours on the documents. According to Table~\ref{tab:ablation_large_supp} (middle), with the exception of WIT, replacing the recurrent cell with cross-attention on the query side seriously degrades performance. Doing that on the document side results in a noticeable improvement on InfoSeek but, in general, we still observe a gap in favor of the original \ours model, where two different recurrent cells independently work on both sides.

\tit{Effect of Text Encoder}
We also train a version of \ours where we replace the CLIP textual encoder with ColBERTv2~\cite{santhanam2022colbertv2}, which has a larger context length of up to 512 tokens, compared to the 77 tokens of CLIP. As shown in Table~\ref{tab:ablation_large_supp} (bottom), this choice translates into a minor improvement of $+0.5$ R@10 points on WIT, and $+1.1$ PR@5 points on E-VQA. However, by sticking with the CLIP textual encoder, \ours enjoys more robust performance across the entire M2KR benchmark, picking with a gain of $+14.9$ and $+7.0$ R@5 points on KVQA and InfoSeek.

\section{Qualitative Results and Visualizations} 
\begin{figure*}[t]
    \centering
    \resizebox{\linewidth}{!}{
    \setlength{\tabcolsep}{.5em}
    \begin{tabular}{p{4.2cm} p{4.2cm} p{4.2cm} p{4.2cm} p{0.1mm}}
    \toprule
    \multicolumn{2}{c}{\textbf{WIT}} & \multicolumn{2}{c}{\textbf{IGLUE}} \\
    \midrule
    \centering\includegraphics[width=0.9\linewidth,height=0.7\linewidth]{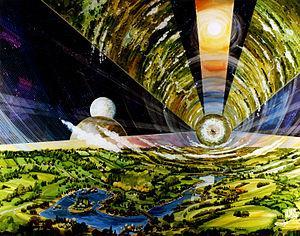} &
    \centering\includegraphics[width=0.9\linewidth,height=0.7\linewidth]{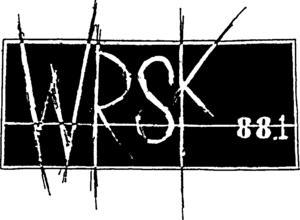} & 
    \centering\includegraphics[width=0.9\linewidth,height=0.7\linewidth]{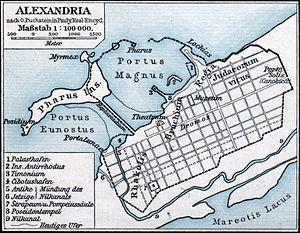} &
    \centering\includegraphics[width=0.9\linewidth,height=0.7\linewidth]{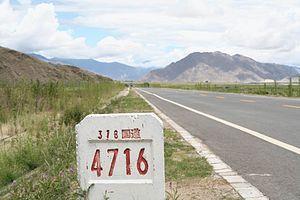} & \\
    \textit{\footnotesize Could you elucidate the document associated with this image?} &  \textit{\footnotesize Provide information about the document linked to this image.} &  \textit{\footnotesize Identify the document that is connected to this image.} & 
    \textit{\footnotesize What document is represented by this image?} \\
    \midrule
    \footnotesize\textbf{PreFLMR~\cite{lin2024preflmr}}: title: Mel Hunter section title: Science Fiction Illustration hierarchical section title: Mel Hunter Biography [...] & 
    \footnotesize\textbf{PreFLMR~\cite{lin2024preflmr}}: title: University of Nebraska Press hierarchical section title: University of Nebraska Press caption attribution [...] & 
    \footnotesize\textbf{PreFLMR~\cite{lin2024preflmr}}: title: Laodicea in Syria section title: History hierarchical section title: Laodicea in Syria History caption: Map showing [...] & 
    \footnotesize\textbf{PreFLMR~\cite{lin2024preflmr}}: title: China National Highway 318 hierarchical section title: China National Highway 318 caption: China National [...] \\
    \midrule
    \footnotesize\textbf{\colorbox{ourcolor}{\ours (Ours)}}: title: Topopolis hierarchical section title: Topopolis caption: Interior view of an O'Neill cylinder space habitat, similar [...] & 
    \footnotesize\textbf{\colorbox{ourcolor}{\ours (Ours)}}: title: WSRU hierarchical section title: WSRU caption: WRSK's "grunge" logo, 1996, Logo for the former WRSK-FM [...] & 
    \footnotesize\textbf{\colorbox{ourcolor}{\ours (Ours)}}: title: Library of Alexandria section title: Early expansion and organization hierarchical section title: Library of [...] & 
    \footnotesize\textbf{\colorbox{ourcolor}{\ours (Ours)}}: title: China National Highway 318 hierarchical section title: China National Highway 318 caption: China National [...] \\
    \\
    \midrule
    \multicolumn{2}{c}{\textbf{KVQA}} & \multicolumn{2}{c}{\textbf{LLaVA}} \\
    \midrule
    \centering\includegraphics[width=0.9\linewidth,height=0.7\linewidth]{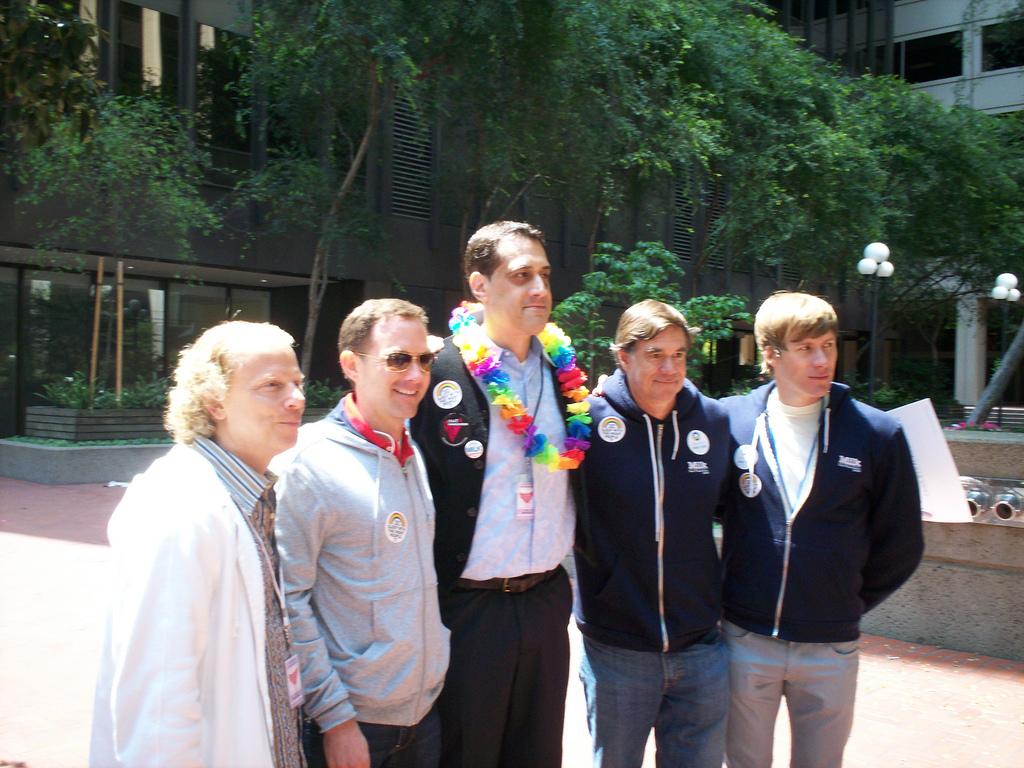} &
    \centering\includegraphics[width=0.9\linewidth,height=0.7\linewidth]{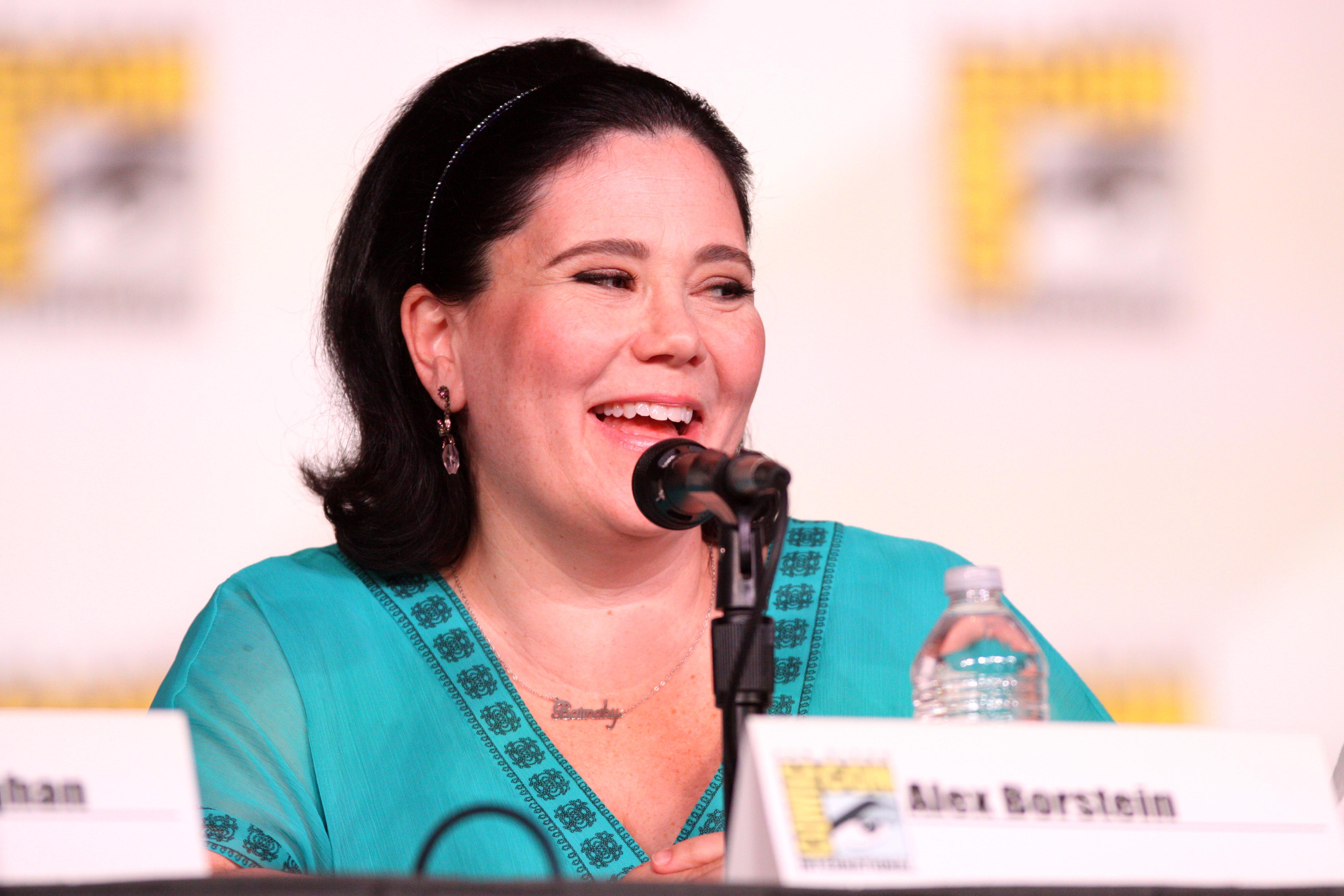} & 
    \centering\includegraphics[width=0.9\linewidth,height=0.7\linewidth]{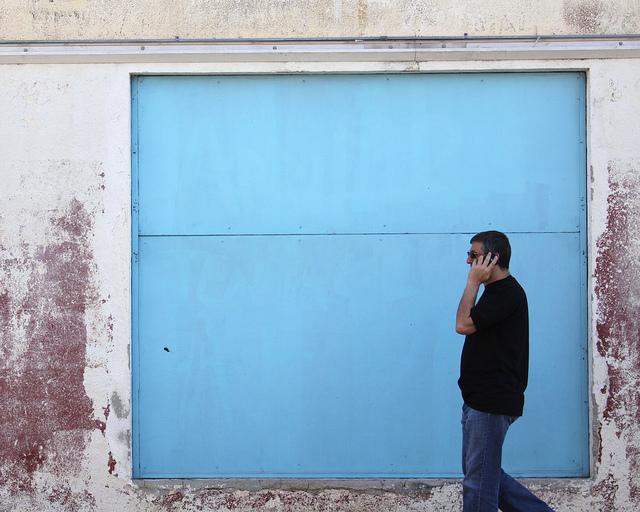} &
    \centering\includegraphics[width=0.9\linewidth,height=0.7\linewidth]{images/qualitatives/M2KR/llava/LLAVA150K_reason_2544.jpg} & \\
    \textit{\footnotesize Provide a brief description of the image and the relevant details of the person in the image.} &  
    \textit{\footnotesize Provide a brief description of the image and the relevant details of the person in the image.} &  
    \textit{\footnotesize What does the background of the image include?} & 
    \textit{\footnotesize What activity might be good for this dog and why?} \\
    \midrule
    \footnotesize\textbf{PreFLMR~\cite{lin2024preflmr}}: This is an image of Lynn Swann and HHS Secretary Tommy Thompson. Tommy Thompson went to University [...] & 
    \footnotesize\textbf{PreFLMR~\cite{lin2024preflmr}}: This is an image of Plumb attending a "Heroes for Autism" event in Hollywood, California, April 2009. Eve Plumb [...] & 
    \footnotesize\textbf{PreFLMR~\cite{lin2024preflmr}}: The background of the image features a wall or a fence that is filled with graffiti art, giving the scene an urban vibe. & 
    \footnotesize\textbf{PreFLMR~\cite{lin2024preflmr}}: The dog's owner is likely engaging in this activity of playing catch with the orange Frisbee to provide physical exercise [...] \\
    \midrule
    \footnotesize\textbf{\colorbox{ourcolor}{\ours (Ours)}}: This is an image of (l. to r.) The two Milk producers Bruce Cohen and Dan Jinks, Stuart, director Gus Van Sant [...] & 
    \footnotesize\textbf{\colorbox{ourcolor}{\ours (Ours)}}: This is an image of Comic-Con 2012. Alex Borstein went to San Francisco State University, date of birth is 1973-02-15 [...] & 
    \footnotesize\textbf{\colorbox{ourcolor}{\ours (Ours)}}: The background of the image includes a blue wall and a reclaimed building in a city. & 
    \footnotesize\textbf{\colorbox{ourcolor}{\ours (Ours)}}: A suitable activity for this dog would be playing Frisbee, as shown in the image. Playing Frisbee not only provides the dog [...] \\
    \bottomrule
    \end{tabular}
    }
    \vspace{-0.15cm}
    \caption{Qualitative results on M2KR, for datasets that do not include document images.}
    \label{fig:qualitatives_m2kr_1}
\end{figure*}

\begin{figure*}[t]
    \centering
    \resizebox{\linewidth}{!}{
    \setlength{\tabcolsep}{.5em}
    \begin{tabular}{p{4.2cm} p{4.2cm} p{4.2cm} p{4.2cm} p{0.1mm}}
    \toprule
    \multicolumn{2}{c}{\textbf{OVEN}} & \multicolumn{2}{c}{\textbf{InfoSeek}} \\
    \midrule
    \centering\includegraphics[width=0.9\linewidth,height=0.7\linewidth]{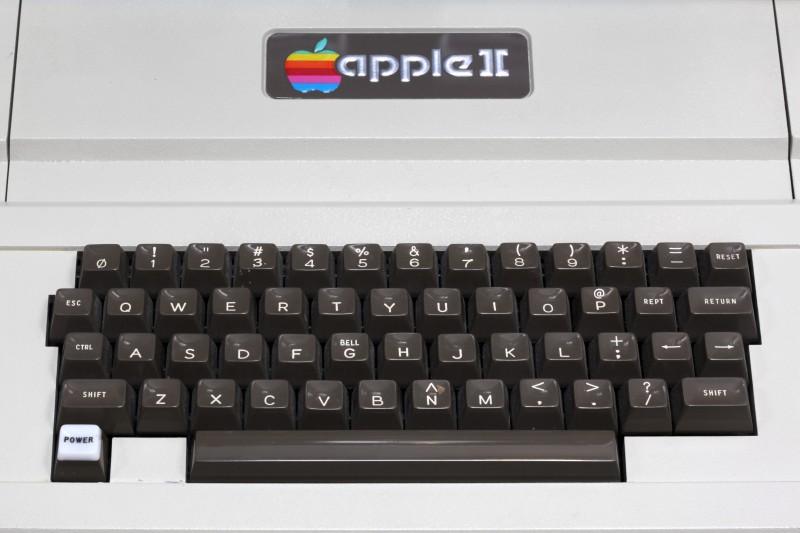} &
    \centering\includegraphics[width=0.9\linewidth,height=0.7\linewidth]{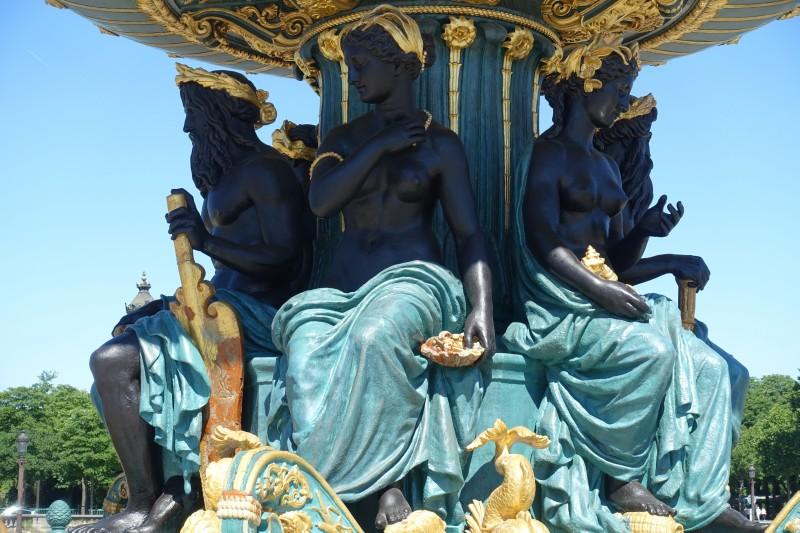} & 
    \centering\includegraphics[width=0.9\linewidth,height=0.7\linewidth]{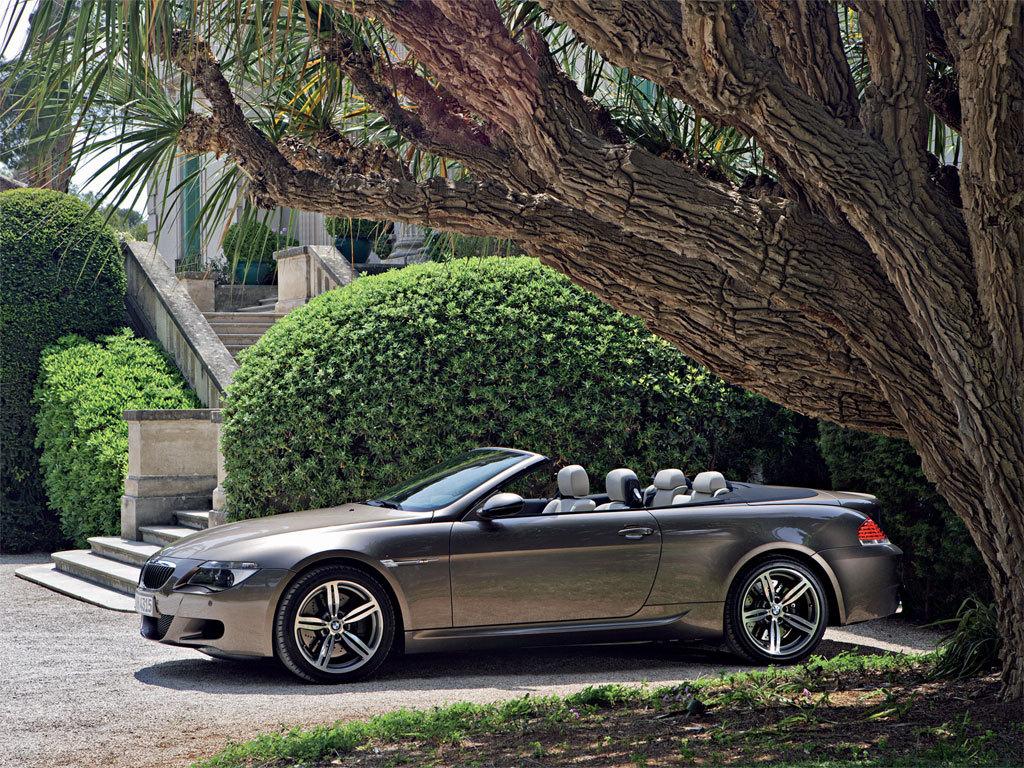} &
    \centering\includegraphics[width=0.9\linewidth,height=0.7\linewidth]{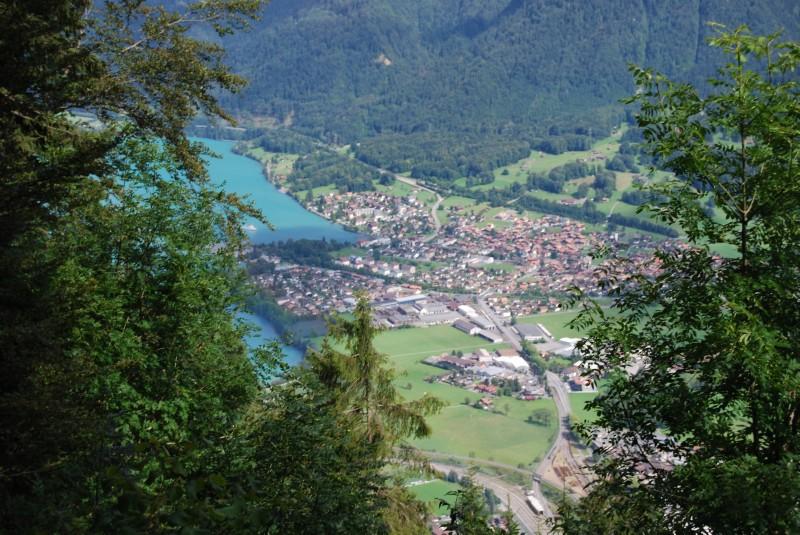} & \\
    \textit{\footnotesize Where is this place?} &  \textit{\footnotesize what is the name of this building?} &  \textit{\footnotesize Which company manufactures this vehicle?} & \textit{\footnotesize Where is the lake outflow to?} \\
    \midrule
    \footnotesize\textbf{PreFLMR~\cite{lin2024preflmr}}: A supercomputer is a computer with a high level of performance as compared to a general-purpose computer. [...] & 
    \footnotesize\textbf{PreFLMR~\cite{lin2024preflmr}}: The Fontaine Louvois is a monumental public fountain in Square Louvois on the rue Richelieu in the Second [...] & 
    \footnotesize\textbf{PreFLMR~\cite{lin2024preflmr}}: [...] On 28 May 2013, Aston Martin announced the V12 Vantage S - a sportier version of the V12 Vantage that [...] & 
    \footnotesize\textbf{PreFLMR~\cite{lin2024preflmr}}: The ships also connect to the Giessbachbahn, a funicular which climbs up to the famous Giessbach Falls. [...] \\
    \midrule
    \footnotesize\textbf{\colorbox{ourcolor}{\ours (Ours)}}: The \textbf{Musée Bolo} or Swiss Museum of Computer Science, Digital Culture and Video Games is a private museum dedicated [...] & 
    \footnotesize\textbf{\colorbox{ourcolor}{\ours (Ours)}}: The \textbf{Fontaines de la Concorde} are two monumental fountains located in the Place de la Concorde in the center of Paris. [...] & 
    \footnotesize\textbf{\colorbox{ourcolor}{\ours (Ours)}}: SMG III gearbox with the E60 M5 that produces at 7,750 rpm and of torque at 6,100 rpm. \textbf{BMW} claimed performance [...] & 
    \footnotesize\textbf{\colorbox{ourcolor}{\ours (Ours)}}: It flows out into a further stretch of the \textbf{Aare} at its western end. The culminating point of the lake's drainage basin is [...] \\
    \addlinespace[1mm]
    \centering\includegraphics[width=0.75\linewidth,height=0.55\linewidth]{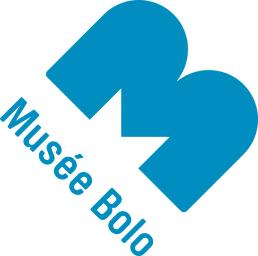} &
    \centering\includegraphics[width=0.75\linewidth,height=0.55\linewidth]{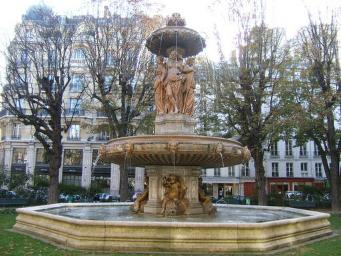}  & 
    \centering\includegraphics[width=0.75\linewidth,height=0.55\linewidth]{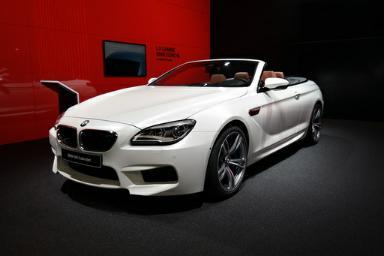} &
    \centering\includegraphics[width=0.75\linewidth,height=0.55\linewidth]{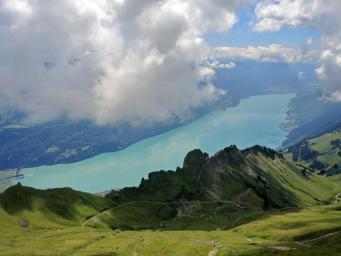} & \\
    \\
    \midrule
    \multicolumn{2}{c}{\textbf{E-VQA}} & \multicolumn{2}{c}{\textbf{OKVQA}} \\
    \midrule
    \centering\includegraphics[width=0.9\linewidth,height=0.7\linewidth]{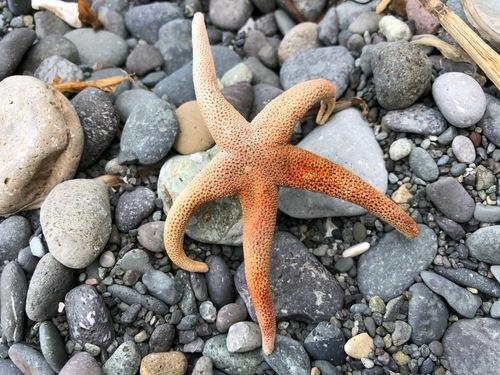} &
    \centering\includegraphics[width=0.9\linewidth,height=0.7\linewidth]{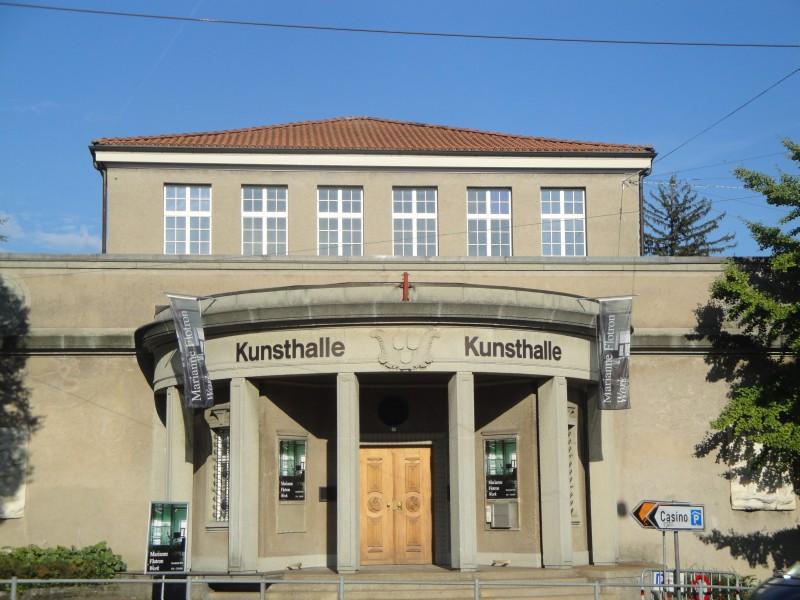} & 
    \centering\includegraphics[width=0.9\linewidth,height=0.7\linewidth]{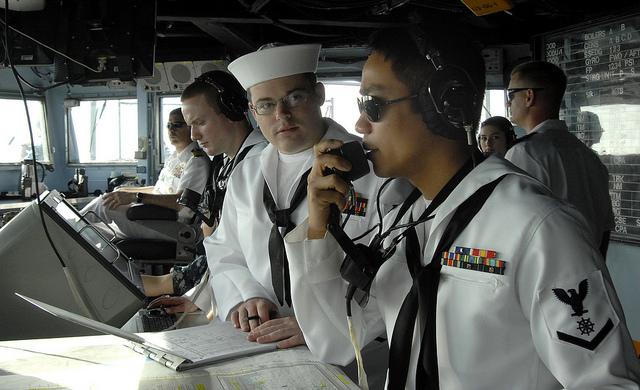} &
    \centering\includegraphics[width=0.9\linewidth,height=0.7\linewidth]{images/qualitatives/M2KR/okvqa/134905.jpg} & \\
    \textit{\footnotesize What does this animal eat?} &  \textit{\footnotesize When was this museum built?} &  \textit{\footnotesize What military branch do the men in the picture belong to?} & \textit{\footnotesize What ingredient is missing from the picture to make a pb and j sandwich?} \\
    \midrule
    \footnotesize\textbf{PreFLMR~\cite{lin2024preflmr}}: At the larval stage, Pisaster ochraceus are filter feeders and their diet consists of plankton. As an adult, P. ochraceus feeds [...] & 
    \footnotesize\textbf{PreFLMR~\cite{lin2024preflmr}}: Built in 1860 by leading local industrialist Wilhelm Vedder, the original building was erected in 1860 with extensive [...] & 
    \footnotesize\textbf{PreFLMR~\cite{lin2024preflmr}}: [...] In modern usage "\textbf{navy}" used alone always denotes a military fleet, although the term "merchant \textbf{navy}" for a [...] & 
    \footnotesize\textbf{PreFLMR~\cite{lin2024preflmr}}: [...] sweet cream, is the highest grade of butter, has a sweet flavor, and is readily spreadable. If the butter is salted, the salt [...] \\
    \midrule
    \footnotesize\textbf{\colorbox{ourcolor}{\ours (Ours)}}: They mainly feed on \textbf{sponges and small bacteria}. The sea star moves these tiny particles, which are captured in mucus and [...] & 
    \footnotesize\textbf{\colorbox{ourcolor}{\ours (Ours)}}: The Kunsthalle Bern is a Kunsthalle (art exposition hall) on the Helvetiaplatz in Bern, Switzerland. It was built in \textbf{1917-1918} [...] & 
    \footnotesize\textbf{\colorbox{ourcolor}{\ours (Ours)}}: A \textbf{navy} or maritime force is the branch of a nation's armed forces principally designated for naval and amphibious warfare [...] & 
    \footnotesize\textbf{\colorbox{ourcolor}{\ours (Ours)}}: be substituted for the \textbf{jelly} component. On the flip side, the popularity of almond butter has inspired some aficionados to [...] \\
    \addlinespace[1mm]
    \centering\includegraphics[width=0.75\linewidth,height=0.55\linewidth]{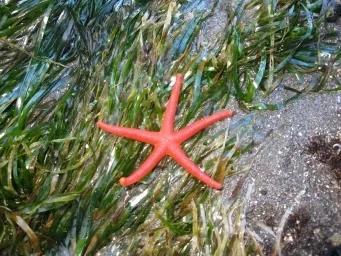} &
    \centering\includegraphics[width=0.75\linewidth,height=0.55\linewidth]{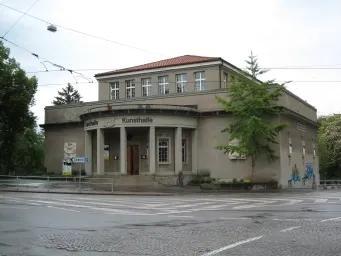}  & 
    \centering\includegraphics[width=0.75\linewidth,height=0.55\linewidth]{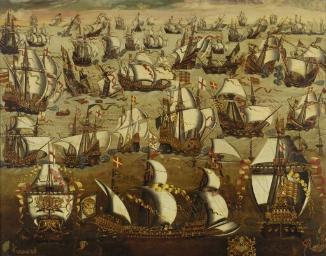} &
    \centering\includegraphics[width=0.75\linewidth,height=0.55\linewidth]{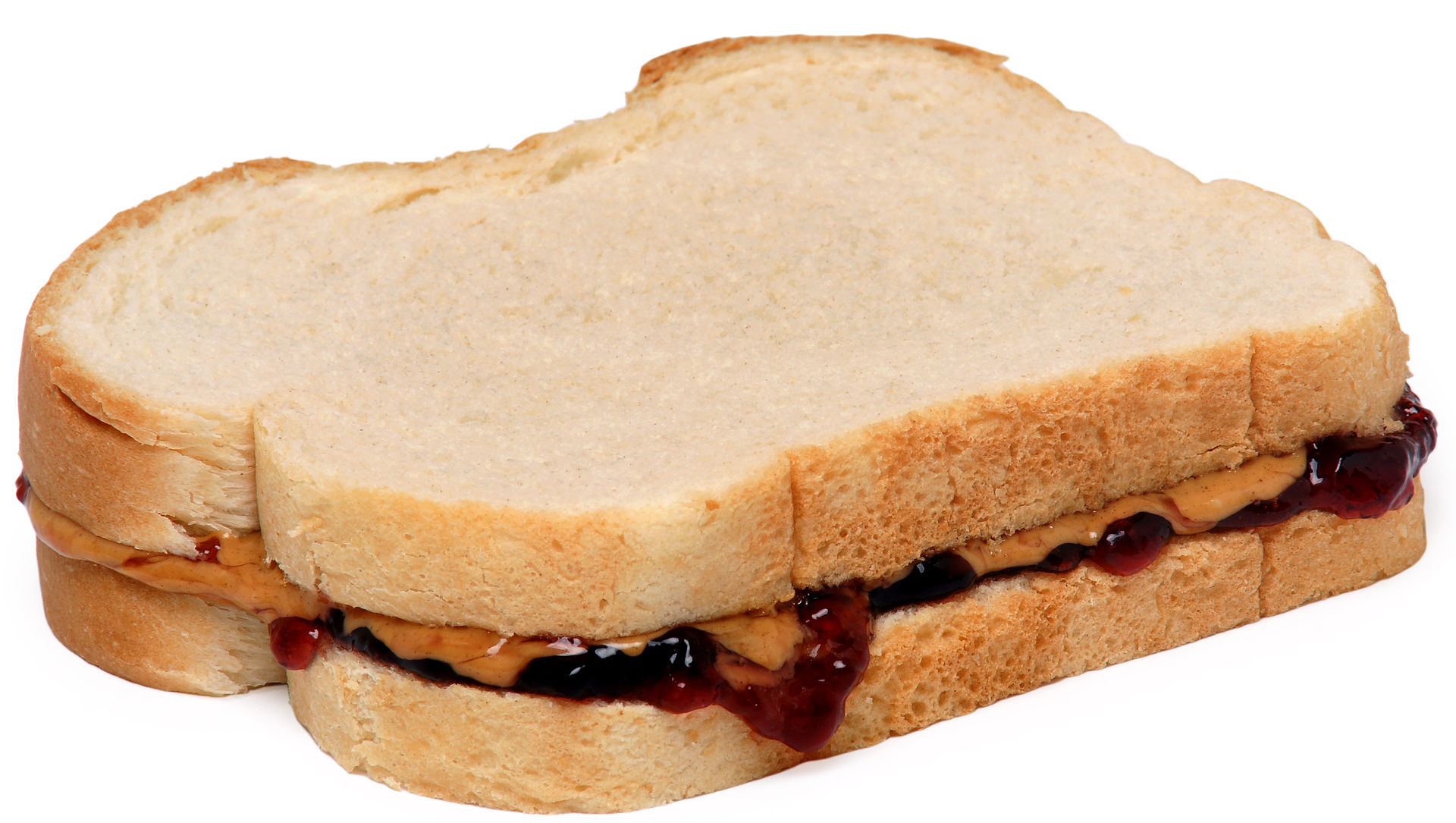} & \\
    \bottomrule
    \end{tabular}
    }
    \vspace{-0.15cm}
    \caption{Qualitative results on M2KR, for datasets that include document images. We highlight the reference answer in bold font whenever it is found in the retrieved text. For \ours, we also add the document image attached to the text.}
    \label{fig:qualitatives_m2kr_2}
\end{figure*}

\tinytit{Results on M2KR} 
In Fig.~\ref{fig:qualitatives_m2kr_1} we present a qualitative comparison between PreFLMR and \ours. To ensure a direct evaluation with PreFLMR, we focus exclusively on datasets from the M2KR benchmark that exclude document images. As it can be seen, \ours consistently retrieves more contextually accurate and detailed information to address the given queries.
In contrast, Fig.~\ref{fig:qualitatives_m2kr_2} reports a qualitative analysis of the difference between PreFLMR and \ours, where the latter can exploit document images, when available. The examples demonstrate that accessing document images significantly improves the ability of the model to answer questions accurately, leveraging the visual context to complement textual information.

\begin{figure*}[t]
    \centering
    \includegraphics[width=0.95\linewidth]{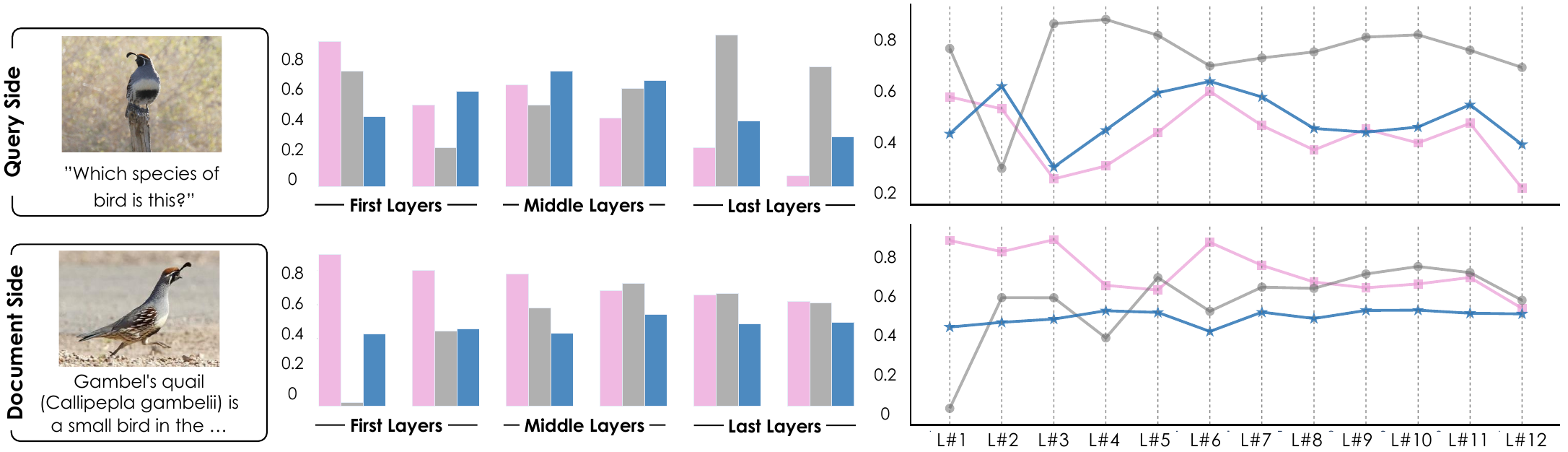}
    \includegraphics[width=0.95\linewidth]{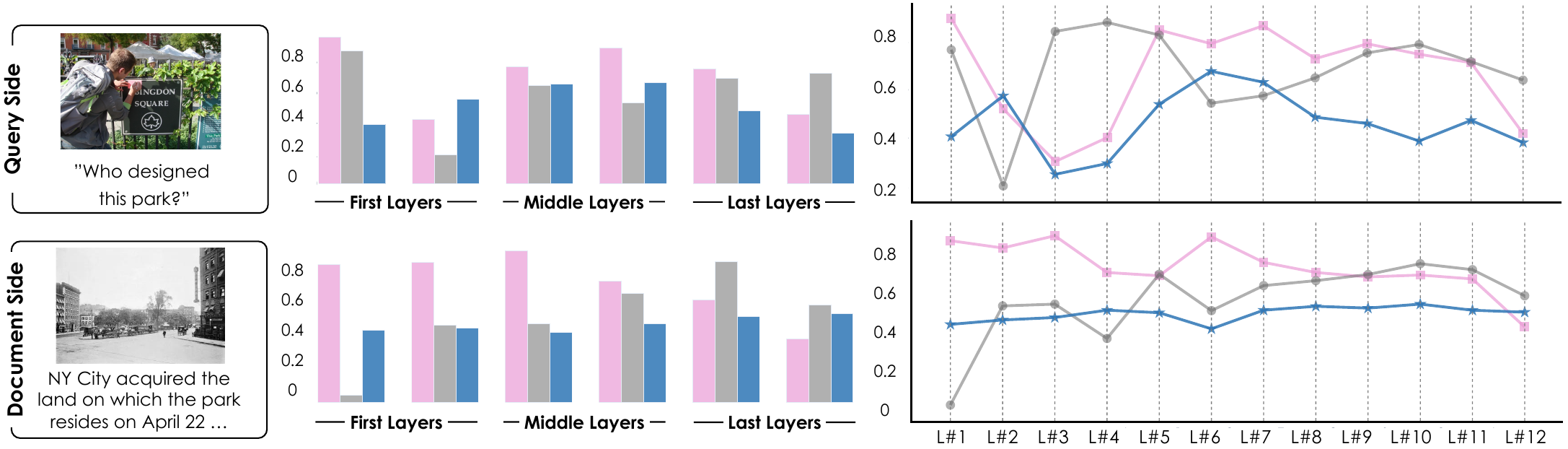}
    \includegraphics[width=0.95\linewidth]{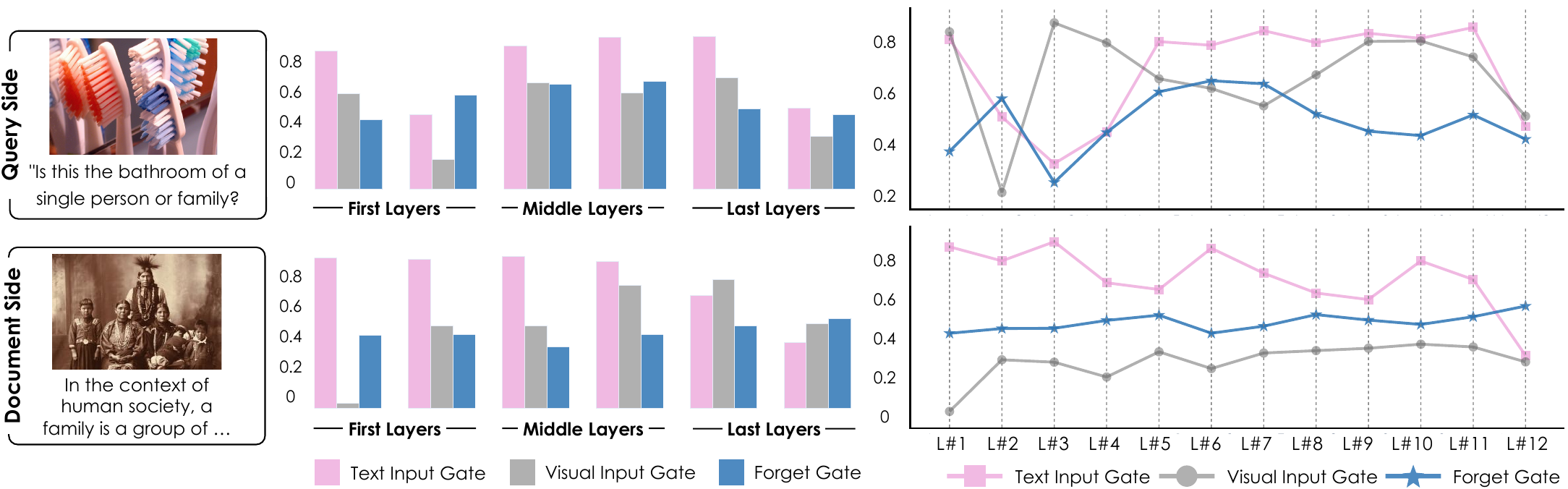}
    \vspace{-0.2cm}
    \caption{Qualitative analysis of gate activations. The left-side displays gate behavior during the encoding of a single query-document pair, while the right shows average activations over 2k examples from OVEN (top), E-VQA (middle), and OKVQA (bottom).}
    \label{fig:gate_activations_supp}
    \vspace{0.2cm}
\end{figure*}

\tit{Qualitative Analysis of Gate Activations} In Fig.~\ref{fig:gate_activations_supp} we show additional activation patterns of our learnable sigmoid gates within the recurrent cell of \ours. On the left, we show gate behavior for a single query-document encoding, while on the right, we show average activations over 2k samples from the OVEN (top), E-VQA (middle), and OKVQA (bottom) split of M2KR. When encoding queries, the behavior of the gates shows a balanced contribution from both the visual and textual components, with the forget gate never dropping to zero, indicating that both modalities consistently provide useful information. Interestingly, the impact of each layer varies significantly, with noticeable fluctuations in gate values across the layers, highlighting the dynamic interaction between modalities throughout the encoding process. Notably, on the OVEN dataset, the contribution from visual information is higher than that from textual input compared to the other two datasets, suggesting a greater reliance on visual input in this setting.

On the document side, overall, the model initially relies more heavily on the textual component in the first layers, with gate activations becoming more stable and balanced in later layers. Notably, in the OKVQA dataset, the gap between activation values is larger compared to the other two datasets. This is especially evident in the single example on the left side, where the image document provides little contribution to answering the question: consequently, the visual input gate maintains low scores across all layers, while the textual information receives greater emphasis.

\begin{figure*}[t]
\begin{minipage}{0.325\linewidth}
\scriptsize{\textbf{Q:} What is this place named after? }\\
\begin{minipage}{0.443\linewidth}
\includegraphics[width=1.\linewidth,height=0.88\linewidth]{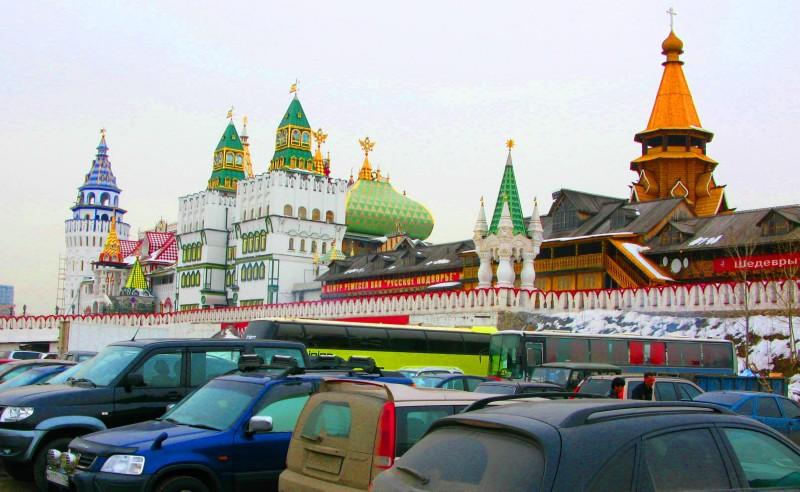}
\end{minipage}
\hfill
\begin{minipage}{0.53\linewidth}
\scriptsize{
\textbf{CLIP~\cite{radford2021learning}}:\\
Moscow \textcolor{red}{\xmark} \\
\textbf{PreFLMR~\cite{lin2024preflmr}}:\\
Golden Gate \textcolor{red}{\xmark} \\
\textbf{\ours (Ours):}\\
Izmaylovo \textcolor[HTML]{00b050}{\cmark}
}
\end{minipage}
\end{minipage}
\hspace{0.02cm}
\begin{minipage}{0.325\linewidth}
\scriptsize{\textbf{Q:} Which city or region does this place locate in? }\\
\begin{minipage}{0.443\linewidth}
\includegraphics[width=1.\linewidth,height=0.88\linewidth]{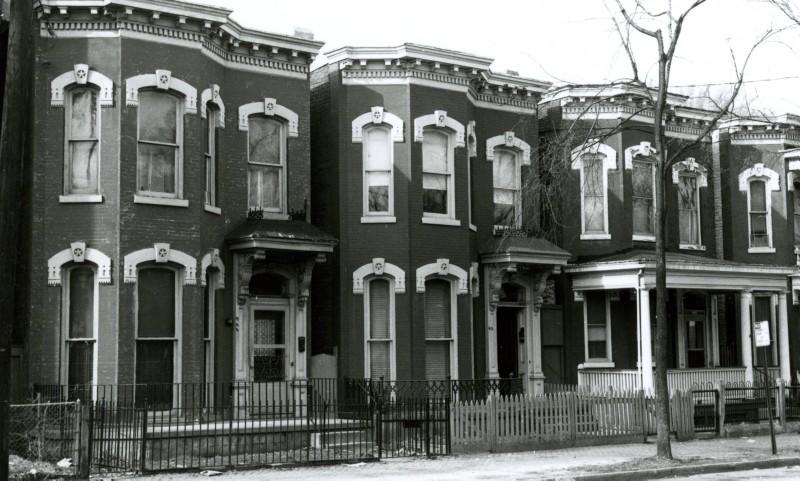}
\end{minipage}
\hfill
\begin{minipage}{0.53\linewidth}
\scriptsize{
\textbf{CLIP~\cite{radford2021learning}}:\\
Bloomington, Illinois \textcolor{red}{\xmark} \\
\textbf{PreFLMR~\cite{lin2024preflmr}}:\\
Illinois \textcolor{red}{\xmark} \\
\textbf{\ours (Ours):}\\
Richmond \textcolor[HTML]{00b050}{\cmark}
}
\end{minipage}
\end{minipage}
\hspace{0.02cm}
\vspace{0.1cm}
\begin{minipage}{0.325\linewidth}
\scriptsize{\textbf{Q:} Who is the founder of the organization in the image? }\\
\begin{minipage}{0.443\linewidth}
\includegraphics[width=1.\linewidth,height=0.88\linewidth]{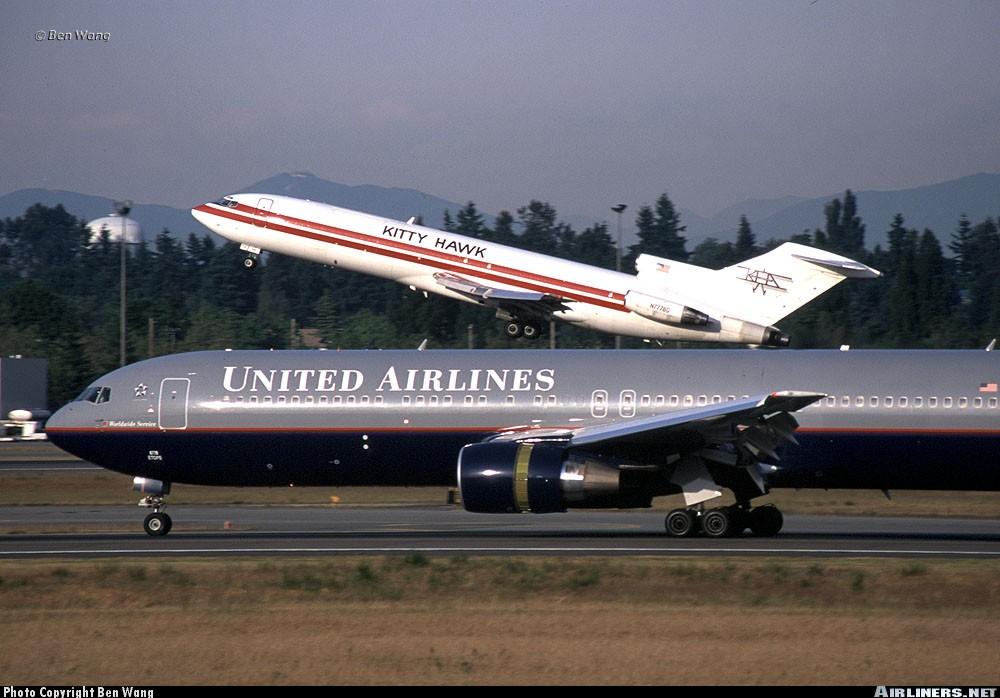}
\end{minipage}
\hfill
\begin{minipage}{0.53\linewidth}
\scriptsize{
\textbf{CLIP~\cite{radford2021learning}}:\\
Madeleine Paulson \textcolor{red}{\xmark} \\
\textbf{PreFLMR~\cite{lin2024preflmr}}:\\
William McPherson Allen \textcolor{red}{\xmark} \\
\textbf{\ours (Ours):}\\
William Boeing \textcolor[HTML]{00b050}{\cmark}
}
\end{minipage}
\end{minipage}
\begin{minipage}{0.325\linewidth}
\scriptsize{\textbf{Q:} What is the closest parent taxonomy of this plant? }\\
\begin{minipage}{0.443\linewidth}
\includegraphics[width=1.\linewidth,height=0.88\linewidth]{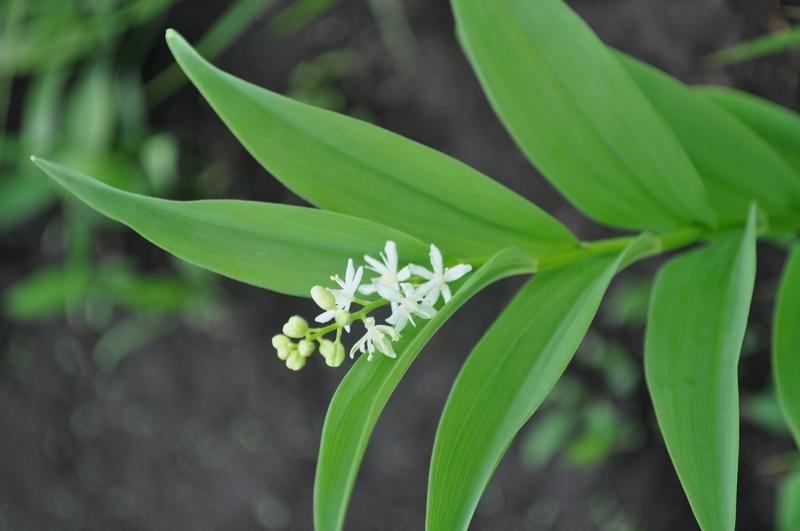}
\end{minipage}
\hfill
\begin{minipage}{0.53\linewidth}
\scriptsize{
\textbf{CLIP~\cite{radford2021learning}}:\\
Allium \textcolor{red}{\xmark} \\
\textbf{PreFLMR~\cite{lin2024preflmr}}:\\
Allium \textcolor{red}{\xmark} \\
\textbf{\ours (Ours):}\\
Maianthemum \textcolor[HTML]{00b050}{\cmark}
}
\end{minipage}
\end{minipage}
\hspace{0.02cm}
\begin{minipage}{0.325\linewidth}
\scriptsize{\textbf{Q:} How many centimetre is the wingspan of this bird? }\\
\begin{minipage}{0.443\linewidth}
\includegraphics[width=1.\linewidth,height=0.88\linewidth]{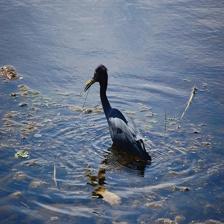}
\end{minipage}
\hfill
\begin{minipage}{0.53\linewidth}
\scriptsize{
\textbf{CLIP~\cite{radford2021learning}}:\\
20 \textcolor{red}{\xmark} \\
\textbf{PreFLMR~\cite{lin2024preflmr}}:\\
144-155 \textcolor{red}{\xmark} \\
\textbf{\ours (Ours):}\\
100-105 \textcolor[HTML]{00b050}{\cmark}
}
\end{minipage}
\end{minipage}
\hspace{0.02cm}
\vspace{0.1cm}
\begin{minipage}{0.325\linewidth}
\scriptsize{\textbf{Q:} What is the length of this bridge in metre? }\\
\begin{minipage}{0.443\linewidth}
\includegraphics[width=1.\linewidth,height=0.88\linewidth]{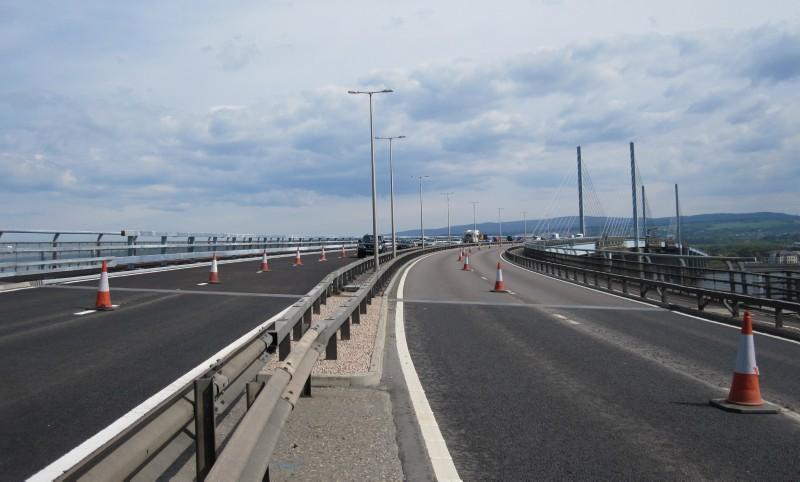}
\end{minipage}
\hfill
\begin{minipage}{0.53\linewidth}
\scriptsize{
\textbf{CLIP~\cite{radford2021learning}}:\\
No answer \textcolor{red}{\xmark} \\
\textbf{PreFLMR~\cite{lin2024preflmr}}:\\
178.4 \textcolor{red}{\xmark} \\
\textbf{\ours (Ours):}\\
1056 \textcolor[HTML]{00b050}{\cmark}
}
\end{minipage}
\end{minipage}
\vspace{-0.25cm}
\caption{VQA results on the validation split of InfoSeek, augmenting LLaVA-MORE with different information retrievers.}
\label{fig:qualitatives_vqa}
\vspace{-0.2cm}
\end{figure*}

\tit{Results on Retrieval-Augmented VQA}
Some qualitative results on sample image-question pairs
from InfoSeek are reported in Fig.~\ref{fig:qualitatives_vqa}, comparing the answers generated by augmenting LLaVA-MORE with different information retrieved by \ours against those retrieved by the original CLIP and PreFLMR models. The results highlight that \ours consistently retrieves more accurate and precise documents that help address specific multimodal questions, outperforming the other approaches. This demonstrates the effectiveness of our architecture in enhancing downstream performance.

\end{document}